\documentclass[10pt,journal,compsoc]{IEEEtran}
\ifCLASSOPTIONcompsoc
\usepackage[nocompress]{cite}
\else
\usepackage{cite}
\fi
\ifCLASSINFOpdf
\else
\fi
\usepackage{amsmath}
\usepackage{amssymb}
\usepackage{amsfonts}
\usepackage{algorithm}
\usepackage{algorithmic}
\usepackage{multicol}
\usepackage{multirow}
\usepackage{makecell}

\usepackage{graphicx}
\usepackage{textcomp}
\usepackage{bm}
\usepackage{epsfig}
\usepackage{dirtree}
\usepackage{booktabs}
\usepackage{array}
\usepackage{makecell}

\usepackage{xspace}
\newcommand{\latinphrase}[1]{\textit{#1}}\usepackage{xspace}
\newcommand{\etal}{\latinphrase{et~al.}\xspace}
\newcommand{\ie}{\latinphrase{i.e.}\xspace}
\newcommand{\etals}{\latinphrase{et~al.}'s\xspace}
\newcommand{\eg}{\latinphrase{e.g.}\xspace}
\newcommand{\etc}{\latinphrase{etc.}\xspace}

\newcommand{\argmax}{\operatornamewithlimits{argmax}}

\usepackage{lineno,hyperref}
\modulolinenumbers[5]

\usepackage{balance}

\newcommand{\xucomment}[1]{}

\begin{document}
	\title{A Survey on Deep Semi-supervised Learning}
	
	\author{Xiangli~Yang,
		Zixing~Song,
		Irwin~King,~\IEEEmembership{Fellow,~IEEE,}
		Zenglin~Xu,~\IEEEmembership{Senior~Member,~IEEE}% stops a space
		\IEEEcompsocitemizethanks{\IEEEcompsocthanksitem X. Yang is with the School of Computer Science and Engineering, University of Electronic Science and Technology of China, Chengdu, China\protect\\
			E-mail: xlyang@std.uestc.edu.cn
			\IEEEcompsocthanksitem Z. Xu is with  the School of Computer Science and Technology, Harbin Institute of Technology,	Shenzhen, China, and Peng Cheng Lab, Shenzhen, China.\protect\\
			Email: xuzenglin@hit.edu.cn
			\IEEEcompsocthanksitem I. King and Z. Song are with the Department of Computer Science and Engineering, The Chinese University of Hong Kong, Hong Kong, China\protect\\
			Email: king@cse.cuhk.edu.hk, zxsong@cse.cuhk.edu.hk
			\IEEEcompsocthanksitem Corresponding author: Zenglin~Xu.
		}% <-this % stops an unwanted space
		\thanks{Manuscript received xx xx, xx; revised xx xx, xx.}
	}

	\IEEEtitleabstractindextext{%
		\begin{abstract}
			Deep semi-supervised learning is a fast-growing field with a range of practical applications. This paper provides a comprehensive survey on both fundamentals and recent advances in deep semi-supervised learning methods from perspectives of model design  and  unsupervised loss functions. We first present a taxonomy for deep semi-supervised learning that categorizes existing methods, including deep generative methods, consistency regularization methods, graph-based methods, pseudo-labeling methods, and hybrid methods. Then we provide a comprehensive review of 52 representative methods and offer a detailed comparison of these methods in terms of the type of losses, contributions, and architecture differences. In addition to the progress in the past few years, we further discuss some shortcomings of existing methods and provide some tentative heuristic solutions for solving these open problems.
		\end{abstract}
		
		}

	\maketitle

	\IEEEdisplaynontitleabstractindextext

	% \fi
	\IEEEpeerreviewmaketitle

	\IEEEraisesectionheading{\section{Introduction}\label{sec:introduction}}

	\IEEEPARstart{D}{eep} learning has been an active research field with abundant applications in  pattern recognition \cite{DBLP:conf/socpar/PadmanabhanP16,DBLP:journals/corr/abs-1809-09645}, data mining \cite{DBLP:journals/inffus/ZhangYCL18a}, statistical learning \cite{DBLP:conf/nips/TranPCP017}, computer vision \cite{DBLP:journals/cacm/KrizhevskySH17,DBLP:conf/cvpr/HeZRS16}, natural language processing \cite{DBLP:conf/naacl/DevlinCLT19,DBLP:journals/corr/abs-2003-01200}, \etc 
	It has achieved great successes in both theory and practice \cite{DBLP:books/daglib/0040158, DBLP:journals/nature/LeCunBH15}, especially in supervised learning scenarios, by leveraging a large amount of high-quality labeled data. However, labeled samples are often difficult, expensive, or time-consuming to obtain. The labeling process usually requires experts' efforts, which is one of the major limitations to train an excellent fully-supervised deep neural network. For example, in medical tasks, the measurements are made with expensive machinery, and labels are drawn from a time-consuming analysis of multiple human experts. If only a few labeled samples are available, it is challenging to build a successful learning system. By contrast, the unlabeled data is usually abundant and can be easily or inexpensively obtained. Consequently, it is desirable to leverage a large number of unlabeled data for improving the learning performance given a small number of labeled samples. For this reason, semi-supervised learning (SSL) has been a hot research topic in machine learning in the last decade \cite{DBLP:books/mit/06/ChapelleSZ06,DBLP:series/synthesis/2009Zhu}.
	
	% fig: Deep semi-supervised learning models
	\begin{figure*}[!t]
		\centering
		\includegraphics[width=6.9in]{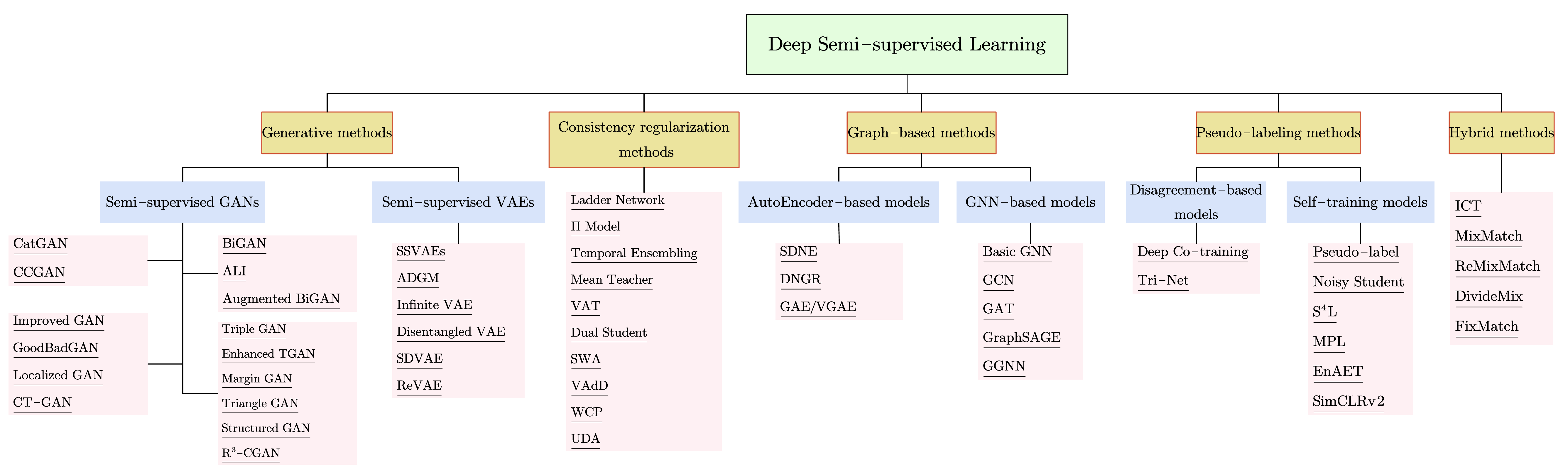}
		\caption{The taxonomy of major deep semi-supervised learning methods based on loss  function and  model design.}
		\label{fig:overview}
	\end{figure*}
	SSL is a learning paradigm associated with constructing models that use both labeled and unlabeled data. SSL methods can improve learning performance by using additional unlabeled instances compared to supervised learning algorithms, which can use only labeled data. It is easy to obtain SSL algorithms by extending supervised learning algorithms or unsupervised learning algorithms.
	SSL algorithms provide a way to explore the latent patterns from unlabeled examples, alleviating the need for a large number of labels \cite{DBLP:conf/nips/OliverORCG18}.  Depending on the key objective function of the systems, one may have a semi-supervised classification, a semi-supervised clustering, or a semi-supervised regression. We provide the definitions as follows:
	\begin{itemize}
		\item \textbf{Semi-supervised classification.} Given a training dataset that consists of both labeled instances and unlabeled instances, semi-supervised classification aims to train a classifier from both the labeled and unlabeled data, such that it is better than the supervised classifier trained only on the labeled data.
		\item \textbf{Semi-supervised clustering.} Given a training dataset that consists of unlabeled instances, and some supervised information about the clusters, the goal of semi-supervised  clustering is to obtain better clustering than the clustering from unlabeled data alone. Semi-supervised clustering is also known as constrained clustering.
		\item \textbf{Semi-supervised regression.} Given a training dataset that consists of both labeled instances and unlabeled instances, the goal of semi-supervised regression is to improve the performance of a regression algorithm from a regression algorithm with  labeled data alone, which predicts a real-valued output instead of a class label.
	\end{itemize}
	In order to explain SSL clearly and concretely, we focus on the problem of image classification. The ideas described in this survey can be adapted without difficulty to other situations, such as object detection, semantic segmentation, clustering, or regression.  Therefore, we primarily review image classification methods with the aid of unlabeled data in this survey.
	
	Since the 1970s when the concept of SSL first came to the fore \cite{DBLP:journals/tit/Agrawala70,DBLP:journals/tit/Fralick67,DBLP:journals/tit/Scudder65a},
	there have been a wide variety of SSL methods, including generative models \cite{DBLP:conf/nips/MillerU96,DBLP:journals/ml/NigamMTM00}, semi-supervised support vector machines \cite{DBLP:conf/icml/Joachims99,DBLP:conf/nips/BennettD98,XuJZKL07nips,XuJZKLY09nips}, graph-based methods \cite{DBLP:conf/icml/ZhuGL03,DBLP:journals/jmlr/BelkinNS06,DBLP:conf/icml/BlumC01,DBLP:conf/nips/ZhouBLWS03}, and co-training \cite{DBLP:conf/colt/BlumM98}. We refer interested readers to \cite{DBLP:books/mit/06/CSZ2006,DBLP:series/synthesis/2009Zhu}, which provide a comprehensive overview of traditional SSL methods.
	Nowadays, deep neural networks have played a dominating role in many research areas.
	It is important to adopt the classic SSL framework and develop novel SSL methods for deep learning settings, which leads to deep semi-supervised learning (DSSL).
	DSSL studies how to effectively utilize both labeled and unlabeled data by deep neural networks. A considerable amount of DSSL methods have been proposed. According to the most distinctive features in  semi-supervised loss functions and model designs, we classify DSSL into five categories, \ie, generative methods, consistency regularization methods, graph-based methods, pseudo-labeling methods, and hybrid methods.
	The overall taxonomy used in this literature is shown in Fig.~\ref{fig:overview}.

	Some representative works of SSL were described in the early survey \cite{DBLP:books/mit/06/CSZ2006,DBLP:series/synthesis/2009Zhu}, however,  emerging technologies based on deep learning, such as adversarial training which generates new training data for SSL, have not been included.
	%For example, Deep semi-supervised methods proposed novel techniques like using adversarial training to generate new training data.
	Besides, \cite{DBLP:conf/nips/OliverORCG18} focuses on unifying the evaluation indices of SSL, and \cite{DBLP:journals/corr/abs-1903-11260} only reviews generative models and teacher-student models in SSL without making a comprehensive overview of SSL. Although \cite{DBLP:journals/ml/EngelenH20} tries to present a whole picture of SSL, the taxonomy is quite different from ours.
	A recent review by Ouali \etal~\cite{Ouali2020AnOO} gives a similar notion of DSSL as we do. However, it does not compare the presented methods based on their taxonomy and provide perspectives on future trends and existing issues. 
	Summarizing both previous and the latest research on SSL, we survey the fundamental theories and compare the deep semi-supervised methods.
	In summary, our contributions are listed as follows.
	\begin{itemize}
		\item We provide a detailed review of DSSL methods and  introduce a taxonomy of major DSSL methods, background knowledge, and various models. One can quickly grasp the frontier ideas of DSSL.
		\item We categorize DSSL methods into several categories, i.e.,  generative methods, consistency regularization methods, graph-based methods, pseudo-labeling methods, and hybrid methods, with particular genres inside each one. 
		We review the variants on each category and give standardized descriptions and unified sketch maps.
		\item We identify several open problems in this field and discuss the future direction for DSSL.
	\end{itemize}
	
	This survey is organized as follows. In Section~\ref{sec:background}, we introduce SSL background knowledge, including assumptions in SSL, classical SSL methods, related concepts, and datasets used in various applications. Section~\ref{sec:regularization} to Section~\ref{sec:hybridModel} introduce the primary deep semi-supervised techniques, \ie, generative methods  in Section~\ref{sec:generative}, consistency regularization methods in Section~\ref{sec:regularization}, graph-based methods  in Section~\ref{sec:graph}, pseudo-labeling methods  in Section~\ref{sec:pseudoLabeling} and hybrid methods in Section~\ref{sec:hybridModel}. In Section~\ref{sec:future_trends}, we discuss the challenges  in semi-supervised  learning and provide some heuristic solutions and future directions for these open problems.

	\section{Background}\label{sec:background}
	Before presenting an overview of the techniques of SSL, we first introduce the notations. To illustrate the DSSL framework, we limit our focus on the single-label classification tasks which are simple to describe and implement. We refer interested readers to \cite{DBLP:conf/cvpr/CevikalpBGS19,DBLP:conf/aaai/WangLQS020,DBLP:journals/pr/CevikalpBG20} for multi-label classification tasks. Let $X = \{X_L,X_U\}$ denote the entire data set, including a small labeled subset $X_L=\{x_i\}_{i=1}^L$ with labels $Y_L=(y_1,y_2,\ldots,y_L)$  and a large scale unlabeled subset $X_U=\{(x_i)\}_{i=1}^U$, and generally we assume $L \ll U$.
	We assume that the dataset contains $K$ classes and the first $L$ examples within $X$ are labeled by $\{y_i\}_{i=1}^L \in (y^1,y^2,\ldots,y^K)$.
	Formally, SSL aims to solve the following optimization problem,
	\begin{equation}
		\min_{\theta} \underset{\text{supervised loss}}{\underbrace{\sum_{x\in X_L,y\in Y_L}\mathcal{L}_s(x,y,\theta)}}+\alpha  \underset{\text{unsupervised loss}}{\underbrace{\sum_{x\in X_U}\mathcal{L}_u(x,\theta)}}+\beta \underset{\text{regularization}}{\underbrace{\sum_{x\in X}\mathcal{R}(x,\theta)}},
		\label{equ: semiLoss}
	\end{equation}
	%%%%%%%%
	\xucomment{
		In the following, we will present an overview of the techniques of SSL. Let $X = \{X_L,X_U\}$ denote the entire data set, including a small labeled subset $X_L=\{(x_i,y_i)\}_{i=1}^L$ with labels $Y_L=(y_1,y_2,\ldots,y_L)$  and a large scale unlabeled subset $X_U=\{(x_i)\}_{i=1}^U$, and $L \ll U$.
		We assume that the dataset contains $K$ classes and the first $L$ examples within $X$ are labeled by $\{y_i\}_{i=1}^L \in (y^1,y^2,\ldots,y^K)$.
		Formally, SSL aims to solve the following optimization problem,
		\begin{equation}
			\min_{\theta} \underset{\text{supervised loss}}{\underbrace{\sum_{(x,y)\in X_L}\mathcal{L}_s(x,y,\theta)}}+\alpha  \underset{\text{unsupervised loss}}{\underbrace{\sum_{x\in X_U}\mathcal{L}_u(x,\theta)}}+\beta \underset{\text{regularization}}{\underbrace{\sum_{x\in X}\mathcal{R}(x,\theta)}},
			\label{equ: semiLoss}
		\end{equation}
	}
	where $\mathcal{L}_s$ denotes the per-example supervised loss, \eg, cross-entropy for classification, $\mathcal{L}_u$ denotes the per-example unsupervised loss, and $\mathcal{R}$ denotes the per-example regularization, \eg, consistency loss or a designed regularization term.  Note that unsupervised loss terms are often not strictly distinguished from regularization terms, as regularization terms are normally not guided by label information. Lastly, $\theta$ denotes the model parameters and $\alpha, \beta \in \mathbb{R}_{>0}$ denotes the trade-off. Different choices of the unsupervised loss functions and regularization terms lead to different semi-supervised models. Note that we do not make a clear distinction between unsupervised loss and regularization terms in many cases. Unless particularly specified, the notations used in this paper are illustrated in TABLE~\ref{tab:notations}. 
	
	Regarding whether test data are wholly available in the training process, semi-supervised learning can be classified into two settings: the transductive setting and the inductive learning setting. Transductive learning assumes that the unlabeled samples in the training process are exactly the data to be predicted, and the purpose of the transductive learning is to generalize over these unlabeled samples, while inductive learning supposes that the learned semi-supervised classifier will be still applicable to new unseen data. In fact, most graph-based methods are transductive while most other kinds of SSL methods are inductive.

	\subsection{Assumptions for semi-supervised learning}\label{subsec:assumptions}
	
	SSL aims to predict more accurately with the aid of unlabeled data than supervised learning that uses only labeled data. However, an essential prerequisite is that the data distribution should be under some assumptions. Otherwise, SSL may not improve supervised learning and may even degrade the prediction accuracy by misleading inferences. Following \cite{DBLP:series/synthesis/2009Zhu} and \cite{DBLP:books/mit/06/CSZ2006}, the related assumptions in SSL include:
	
	\textbf{Self-training assumption.} The predictions of the self-training model, especially those with high confidence, tend to be correct. We can assume that when the hypothesis is satisfied, those high-confidence  predictions are considered to be ground-truth. This can happen when classes form well-separated clusters. 
	
	\textbf{Co-training assumption.} Different reasonable assumptions lead to different combinations of labeled and unlabeled data, and accordingly, different algorithms are designed to take advantage of these  combinations. For example, Blum \etal \cite{DBLP:conf/colt/BlumM98} proposed a co-training model, which works under the assumptions: instance $x$ has two conditionally independent views, and each view is sufficient for a classification task. %However, \cite{DBLP:journals/tkde/ZhouL05} is not constrained  by these assumptions.
	
	\textbf{Generative model assumption.} Generally, it is assumed that data are  generated from a mixture of distributions. When the number of mixed components, a prior $p(y)$ and a conditional distribution $p(x|y)$ are correct,  data can be assumed to come from the mixed model. This assumption suggests that if the generative model is correct enough, we can establish a valid link between the distribution of unlabeled data and the category labels by $p(x,y)=p(y)p(x|y)$.
	
	\xucomment{
		\textbf{Semi-supervised smoothness assumption.}
		If point $x_1$ is close to $x_2$ in a high-density region, then the corresponding outputs $y_1$ and $y_2$ are also close.
		The hypothesis implies that if two samples belong to the same cluster, their outputs will likely be the same class label (see cluster assumption in \cite{DBLP:books/mit/06/CSZ2006}). On the contrary, they are separated by a low-density region, and then their output class labels tend to be different.
	}
	
	%\textbf{Graph-based semi-supervised learning assumption.} If two instances $x_1$ and $x_2$ are connected by a strong edge in graph $\mathcal{G}$,  they have same target labels. This assumption reflects the local smoothness of the labels on the graph, so they vary slowly. 

	%\textbf{Semi-supervised smoothness assumption.} If a data point  $x_1$ is close to $x_2$ in a high-density region, then the corresponding outputs $y_1$ and $y_2$ are also close \cite{DBLP:books/mit/06/CSZ2006}. This assumption implies that if two samples belong to the same cluster, they will likely share the same class label. On the contrary, if they are separated by a low-density region,  then their output class labels tend to be different.
	
	\textbf{Cluster assumption.} If two points $x_1$ and $x_2$ are in the same cluster, they should belong to the same category \cite{DBLP:books/mit/06/CSZ2006}.
	This assumption refers to the fact that data in a single class tend to form a cluster, and when the data points can be connected by short curves that do not pass through any low-density regions, they belong to the same class cluster \cite{DBLP:books/mit/06/CSZ2006}.  According to this assumption, the decision boundary should not cross high-density areas but instead lie in low-density regions \cite{DBLP:conf/aistats/ChapelleZ05}. Therefore, the learning algorithm can use a large amount of  unlabeled data to adjust the classification boundary.
	
	\textbf{Low-density separation.}  The decision boundary should be in a low-density region, not through a high-density area \cite{DBLP:books/mit/06/CSZ2006}. The low-density separation assumption is closely related to the cluster assumption. We can consider the clustering assumption from another perspective by assuming that the class is separated by areas of low density \cite{DBLP:books/mit/06/CSZ2006}. Since the decision boundary in a high-density region would cut a cluster into two different classes and within such a part  would violate the cluster assumption.

	\textbf{Manifold assumption.} If two points $x_1$ and $x_2$ are located in a local neighborhood in the low-dimensional manifold, they have similar class labels \cite{DBLP:books/mit/06/CSZ2006}. This assumption reflects the local smoothness of the decision boundary. It is well known that one of the problems of machine learning algorithms is the curse of dimensionality. It is hard to estimate the actual data distribution when volume grows exponentially with the dimensions in high dimensional spaces. If the data lie on a low-dimensional manifold, the learning algorithms can avoid the curse of dimensionality and operate in the corresponding low-dimension space. %The main difference from the cluster assumption is that the cluster assumption mainly focuses on global distribution, and the manifold assumption primarily considers the locality of the data set.
	
	%table: Notations
	\begin{table}
		\centering
		\caption{Notations}
		\begin{tabular}{c|c}
			\hline
			\textbf{Symbol} & \textbf{Explanation} \\
			\hline
			$\mathcal{X}$ & Input space, for example $\mathcal{X}=\mathbb{R}^n$ \\
			$\mathcal{Y}$ & Output space. \\ & Classification: $\mathcal{Y}=\{y^1,y^2,\ldots, y^K\}$.\\ & Regression: $\mathcal{Y}=\mathbb{R}$  \\
			$X_L$ & Labeled dataset. $x_i\in \mathcal{X}, y_i\in \mathcal{Y}$\\
			$X_U$ &Unlabeled dataset. $x_i\in \mathcal{X}$\\
			$X$ & Input dataset $X$. $N=L+U, L\ll U$\\
			$\mathcal{L}$ & Loss Function \\
			$G$ & Generator \\
			$D$ & Discriminator \\
			$C$ & Classifier \\
			$H$ & Entropy \\
			$\mathbb{E}$ & Expectation \\
			$\mathcal{R}$ & Consistency constraint \\
			$\mathcal{T}_x$& Consistency target \\
			$\mathcal{G}$ & A graph \\
			$\mathcal{V}$ & The set of vertices in a graph \\
			$\mathcal{E}$ & The set of edges in a graph\\
			$v$ & A node $v \in \mathcal{V}$\\
			$e_{ij}$ & An edge linked between node $i$ and $j$, $e_{ij}\in \mathcal{E}$\\
			${A}$ & The adjacency matrix of a graph  \\
			${D}$ & The degree matrix of a graph \\
			$D_{ii}$ & The degree of node $i$\\
			$W$ & The weight matrix\\
			$W_{ij}$ & The weight associated with edge $e_{ij}$\\
			$\mathcal{N}(v)$ & The neighbors of a node $v$\\
			$\mathbf{Z}$ & Embedding matrix \\
			$\mathbf{z}_{v}$ & An embedding for node $v$\\
			$\mathbf{S}$ & Similarity matrix of a graph \\
			$\mathbf{S}[u,v]$ & Similarity measurement between node $u$ and $v$\\
			$\mathbf{h}_{v}^{(k)}$ & Hidden embedding for node $v$ in $k$th layer\\
			$\mathbf{m}_{\mathcal{N}(v)}$ & Message aggregated from node $v$'s neighborhoods \\
			\hline
		\end{tabular}
		\label{tab:notations}
	\end{table}
	
	\subsection{Classical Semi-supervised learning Methods}
	\label{subsec:classical}
	In the following, we briefly introduce some representative SSL methods motivated from the above described assumptions. 
	
	In the 1970s,  the concept of SSL first came to the fore \cite{DBLP:journals/tit/Agrawala70,DBLP:journals/tit/Fralick67,DBLP:journals/tit/Scudder65a}.
	Perhaps the earliest method has been established as self-learning --- an iterative mechanism that uses initial labeled data to train a model to predict some unlabeled samples. Then the most confident prediction is marked as the best prediction of the current supervised model, thus providing more training data for the supervised algorithm, until all the unlabelled examples have been predicted.
	
	Co-training \cite{DBLP:conf/colt/BlumM98} provides a similar solution by training two different models on two different views.
	Confident predictions of one view are then used as labels for the other model. More relevant literature of this method also includes \cite{DBLP:conf/acl/Yarowsky95,DBLP:conf/nips/Sa93,DBLP:conf/ecml/VittautAG02}, \etc
	
	Generative models assume a model $p(x,y)=p(y)p(x|y)$, where the density function $p(x|y)$ is an identifiable distribution, for example, polynomial, Gaussian mixture distribution, \etc, and the uncertainty is the parameters of $p(x|y)$. Generative models can be optimized by using iterative algorithms. \cite{DBLP:conf/nips/MillerU96,DBLP:journals/ml/NigamMTM00} apply EM algorithm for classification. They compute the parameters of $p(x|y)$ and then classify unlabeled instances according to the Bayesian full probability formula. Moreover, generative models are harsh on some assumptions. Once the hypothetical $p(x|y)$ is poorly matched with the actual distribution, it can lead to classifier performance degradation.

	A representative example following the low-density separation principle is Transductive Support Vector Machines (TSVMs) \cite{DBLP:conf/icml/Joachims99,DBLP:conf/nips/BennettD98,DBLP:conf/aistats/ChapelleZ05,DBLP:conf/icml/ChapelleCZ06, DBLP:conf/nips/XuJZKL07}. 
	%An alternative technology that became famous during the ``Support Vector Machine (SVM) boom" of the late 1990's was semi-supervised support vector machines \cite{DBLP:conf/icml/Joachims99,DBLP:conf/nips/BennettD98,DBLP:conf/aistats/ChapelleZ05,DBLP:conf/icml/ChapelleCZ06,DBLP:conf/nips/XuJZKL07}. 
	%S3VMs can also derive 
	As regular SVMs,   TSVMs optimize the gap between decision boundaries and data points, and then expand this gap based on the distance from unlabeled data to the decision margin. To address the corresponding non-convex optimization problem, a number of optimization algorithms have been proposed. For instance,
	in \cite{DBLP:conf/aistats/ChapelleZ05}, a smooth loss function substitutes the hinge loss of the TSVM, and for the decision boundary in a low-density space, a gradient descent technique may be used. %And \cite{DBLP:conf/icml/ChapelleCZ06} extends TSVM by utilizing an iterative method, starting with minimizing a simple convex objective function. Then more complicated procedures are used for the eventual approximation of TSVM's  objective function.

	Graph-based methods rely on the geometry of the data induced by both labeled and unlabeled examples. This geometry is represented by an empirical graph $\mathcal{G}=(\mathcal{V, E})$, where nodes $\mathcal{V}$ represent the training data points with $|\mathcal{V}|=n$ and edges $\mathcal{E}$ represent similarities between the points. By exploiting the graph or manifold structure of data, it is possible to learn with very few labels to propagate information through the graph \cite{Zhu2002LearningFL,DBLP:conf/icml/ZhuGL03,DBLP:books/mit/06/BengioDR06,DBLP:conf/icml/BlumC01,DBLP:conf/nips/ZhouBLWS03,DBLP:journals/jmlr/BelkinNS06}. For example, Label propagation \cite{Zhu2002LearningFL} is to predict the label information of unlabeled nodes from labeled nodes. Each node label propagates to its neighbors according to the similarity.At each step of node propagation, each node updates its label according to its neighbors' label information. In the label propagation label, the label of the labeled data is fixed so that it propagates the label to the unlabeled data. The label propagation method can be applied to deep learning \cite{DBLP:series/lncs/WestonRMC12}.

	\subsection{Related Learning Paradigms} \label{sec:relatedWork}
	There are many learning paradigms that can make use of an extra data source to boost learning performance. 
	Based on the availability of labels or the distribution difference in the extra data source, 
	there are several learning paradigms related to semi-supervised learning. 
	
	\textbf{Transfer learning.}
	Transfer learning \cite{DBLP:journals/tkde/PanY10,DBLP:conf/icann/TanSKZYL18,DBLP:journals/pieee/ZhuangQDXZZXH21} aims to apply knowledge from one or more source domains to a target domain in order to improve performance on the target task. In contrast to SSL, which works well under the hypothesis that the training set and the testing set are independent and identically distributed (i.i.d.), transfer learning allows domains, tasks, and distributions used in training and testing be different but related.
	
	\textbf{Weakly-supervised learning.}
	Weakly-supervised learning \cite{Li2019TowardsSW} relaxes the data dependence that requires ground-truth labels to be given for a large amount of training data set in strong supervision. There are three types of weakly supervised data: incomplete supervised data, inexact supervised data, and inaccurate supervised data. Incomplete supervised data means only a subset of training data is labeled. In this case, representative approaches are SSL and domain adaptation. Inexact supervised data suggests that the labels of training examples are coarse-grained, \eg, in the scenario of multi-instance learning. Inaccurate supervised data means that the given labels are not always ground-truth, such as in the situation of label noise learning.
	
	\textbf{Positive and unlabeled learning.}
	Positive and unlabeled (PU) learning \cite{DBLP:conf/iisa/JaskieS19, DBLP:journals/ml/BekkerD20} is a variant of positive  and negative binary classification, where the training data consists of positive samples and unlabeled samples. Each unlabeled instance can be either the positive and negative class. During the training procedure, only positive samples and unlabeled samples are available. We can think of PU learning as a special case of SSL.
	
	\textbf{Meta-learning.}
	Meta-learning \cite{DBLP:journals/corr/abs-2004-05439, DBLP:journals/corr/abs-2004-11149, DBLP:journals/corr/abs-1810-03548, DBLP:journals/corr/abs-2007-09604,DBLP:journals/corr/abs-2010-03522}, also known as ``learning to learn", aims to learn new skills or adapt to new tasks rapidly with previous knowledge and a few training examples. It is well known that a good machine learning model often requires a large number of samples for training. The meta-learning model is expected to adapt and generalize to new environments that have been encountered during the training process. The adaptation process is essentially a mini learning session that occurs during the test but has limited exposure to new task configurations. Eventually, the adapted model can be trained on various learning tasks and optimized on the distribution of functions, including potentially unseen tasks.
	
	\textbf{Self-supervised learning.}
	Self-supervised learning \cite{DBLP:journals/corr/abs-1902-06162, DBLP:journals/corr/abs-2006-08218,DBLP:journals/corr/abs-2011-00362} has gained popularity due to its ability to prevent the expense of annotating large-scale datasets. It can leverage input data as supervision and use the learned feature representations for many downstream tasks. In this sense, self-supervised learning meets our expectations for efficient learning systems with fewer labels, fewer samples, or fewer trials. Since there is no manual label involved, self-supervised learning can be regarded as a branch of unsupervised learning.
	
\subsection{Datasets and applications}\label{sec:applications}
%\section{Datasets and applications}
SSL has many applications across different tasks and domains, such as image classification, object detection, semantic segmentation, \etc in the domain of computer vision, and text classification, sequence learning, \etc in the field of Natural Language Processing (NLP).
As shown in TABLE~\ref{tab:application}, we summarize some of the most widely used datasets and representative references according to the applications area. In this survey, we mainly discuss the methods applied to image classification since these methods can be extended to other applications, such as \cite{DBLP:conf/nips/JeongLKK19} applied consistency training to object detection, \cite{DBLP:conf/iccv/SoulySS17} modified Semi-GANs to fit the scenario of semantic segmentation. There are many works that have achieved state-of-the-art performance within different  applications. Most of these methods share details of the implementation and source code, and we refer the interested readers to a more detailed review of the datasets and corresponding references in TABLE~\ref{tab:application}.

\begin{table*}
	\caption{Summary of Applications and Datasets}
	\label{tab:application}
	\centering
	%\begin{tabular}{>{\raggedright\arraybackslash}m{3.5cm}|>{\raggedright\arraybackslash}m{9cm}|>{\raggedright\arraybackslash}m{3cm}}
	\begin{tabular}{>{\centering\arraybackslash}m{3.5cm}|>{\centering\arraybackslash}m{9cm}|>{\centering\arraybackslash}m{3cm}}
		\hline
		\textbf{Applications} & \textbf{Datasets} & \textbf{Citations} \\
		\hline
		Image classification &
		MNIST \cite{LeCun2005TheMD},
		SVHN \cite{Netzer2011ReadingDI},
		STL-10 \cite{DBLP:journals/jmlr/CoatesNL11},
		Cifar-10,Cifar-100 \cite{Krizhevsky2009LearningML},
		ImageNet \cite{DBLP:conf/nips/KrizhevskySH12}
		&\cite{DBLP:conf/nips/RasmusBHVR15,DBLP:conf/nips/SajjadiJT16,DBLP:conf/iclr/LaineA17,DBLP:conf/nips/TarvainenV17,DBLP:conf/cvpr/ZhangLH20WCP,DBLP:conf/ijcai/VermaLKBL19,DBLP:conf/nips/BerthelotCGPOR19,DBLP:conf/iclr/BerthelotCCKSZR20,DBLP:conf/iclr/LiSH20}  \\
		\hline
		Object detection & PASCAL VOC \cite{DBLP:journals/ijcv/EveringhamGWWZ10},
		MSCOCO \cite{DBLP:conf/eccv/LinMBHPRDZ14},
		ILSVRC \cite{DBLP:journals/ijcv/RussakovskyDSKS15} &\cite{DBLP:conf/cvpr/TangWGDGC16,DBLP:conf/nips/JeongLKK19,DBLP:conf/iccv/GaoWDLN19,DBLP:journals/corr/abs-2005-04757}  \\
		\hline
		3D object detection & SUN RGB-D \cite{DBLP:conf/cvpr/SongLX15},
		ScanNet \cite{DBLP:conf/cvpr/DaiCSHFN17},
		KITTI \cite{DBLP:conf/cvpr/GeigerLU12} & \cite{DBLP:conf/cvpr/ZhaoCL20,DBLP:conf/iccv/TangL19} \\
		\hline
		Video salient object detection &VOS \cite{DBLP:journals/tip/LiXC18},
		DAVIS \cite{DBLP:conf/cvpr/PerazziPMGGS16},
		FBMS \cite{DBLP:conf/eccv/BroxM10} &\cite{DBLP:conf/iccv/YanL0LWCL19}\\
		\hline
		Semantic segmentation &PASCAL VOC \cite{DBLP:journals/ijcv/EveringhamGWWZ10},
		PASCAL context \cite{DBLP:conf/cvpr/MottaghiCLCLFUY14},
		MS COCO \cite{DBLP:conf/eccv/LinMBHPRDZ14},
		Cityscapes \cite{DBLP:conf/cvpr/CordtsORREBFRS16},
		CamVid \cite{DBLP:conf/eccv/BrostowSFC08},
		SiftFlow \cite{DBLP:journals/pami/LiuYT11,DBLP:conf/cvpr/XiaoHEOT10},
		StanfordBG \cite{DBLP:conf/iccv/GouldFK09}
		&
		\cite{DBLP:conf/iccv/DaiHS15,DBLP:conf/nips/HongNH15,DBLP:conf/iccv/PapandreouCMY15,DBLP:conf/iccv/SoulySS17,
			DBLP:conf/cvpr/WeiXSJFH18,DBLP:conf/cvpr/LeeKLLY19,DBLP:journals/corr/abs-1908-05724,DBLP:conf/iccv/KalluriVCJ19,DBLP:conf/cvpr/IbrahimVRM20,DBLP:conf/cvpr/OualiHT20}\\
		\hline
		Text classification & AG News \cite{DBLP:conf/nips/ZhangZL15},
		BPpedia \cite{lrec12mendes2},
		Yahoo! Answers\cite{DBLP:conf/aaai/ChangRRS08},
		IMDB \cite{DBLP:conf/acl/MaasDPHNP11},
		Yelp review \cite{DBLP:conf/nips/ZhangZL15},
		Snippets \cite{DBLP:conf/www/PhanNH08},
		Ohsumed \cite{DBLP:conf/aaai/YaoM019},
		TagMyNews \cite{DBLP:conf/ecir/VitaleFS12},
		MR \cite{DBLP:conf/acl/PangL05},
		Elec \cite{DBLP:conf/nips/JohnsonZ15},
		Rotten Tomatoes \cite{DBLP:conf/acl/PangL05},
		RCV1 \cite{DBLP:journals/jmlr/LewisYRL04} &\cite{DBLP:conf/aaai/XuSDT17,DBLP:conf/iclr/MiyatoDG17,DBLP:conf/aaai/SachanZS19,DBLP:conf/emnlp/HuYSJL19,DBLP:conf/emnlp/JoC19,DBLP:conf/acl/ChenYY20}  \\
		\hline
		Dialogue policy learning &MultiWOZ \cite{DBLP:conf/emnlp/BudzianowskiWTC18}  &\cite{DBLP:conf/acl/HuangQSZ20}  \\
		\hline
		Dialogue generation &Cornell Movie Dialogs Corpus \cite{DBLP:conf/acl-cmcl/Danescu-Niculescu-Mizil11},
		Ubuntu Dialogue Corpus \cite{DBLP:conf/sigdial/LowePSP15}  &\cite{DBLP:conf/emnlp/ChangHWZYW19}  \\
		\hline
		Sequence learning & IMDB \cite{DBLP:conf/acl/MaasDPHNP11},
		Rotten Tomatoes \cite{DBLP:conf/acl/PangL05},
		20 Newsgroups \cite{DBLP:conf/icml/Lang95},
		DBpedia \cite{lrec12mendes2},
		CoNLL 2003 NER task \cite{DBLP:conf/conll/SangM03},
		CoNLL 2000 Chunking task \cite{DBLP:conf/conll/SangB00},
		Twitter POS Dataset \cite{DBLP:conf/acl/GimpelSODMEHYFS11,DBLP:conf/naacl/OwoputiODGSS13},
		Universal Dependencies(UD) \cite{DBLP:conf/acl/McDonaldNQGDGHPZTBCL13},
		Combinatory Categorial Grammar (CCG) supertagging \cite{DBLP:journals/coling/HockenmaierS07}
		&\cite{DBLP:conf/nips/DaiL15,DBLP:conf/acl/PetersABP17,DBLP:conf/acl/Rei17,DBLP:conf/emnlp/ClarkLML18}  \\
		\hline
		Semantic role labeling& CoNLL-2009 \cite{DBLP:conf/conll/HajicCJKMMMNPSSSXZ09},
		CoNLL-2013 \cite{DBLP:conf/conll/PradhanMXNBUZZ13} &\cite{DBLP:journals/jidm/CarneiroCZR17,DBLP:conf/emnlp/MehtaLC18,DBLP:conf/emnlp/CaiL19}  \\
		\hline
		Question answering	&SQuAD \cite{DBLP:conf/emnlp/RajpurkarZLL16},
		TriviaQA \cite{DBLP:conf/acl/JoshiCWZ17} &\cite{DBLP:conf/aaai/OhTHITK16,DBLP:conf/naacl/DhingraPR18,DBLP:conf/emnlp/ZhangB19}  \\
		\hline
	\end{tabular}
\end{table*}

	\section{Generative methods}
	\label{sec:generative}
	As discussed in Subsection~\ref{subsec:classical}, generative methods can learn the implicit features of data to better model data distributions. They model the real data distribution from the training dataset and then generate new data with this distribution. In this section, we review the deep generative semi-supervised methods based on the Generative Adversarial Network (GAN) framework and the Variational AutoEncoder (VAE) framework, respectively.
	
	%fig: semi-supervised GAN models
	\begin{figure*}[!t]
		\centering
		\includegraphics[width=7.0in]{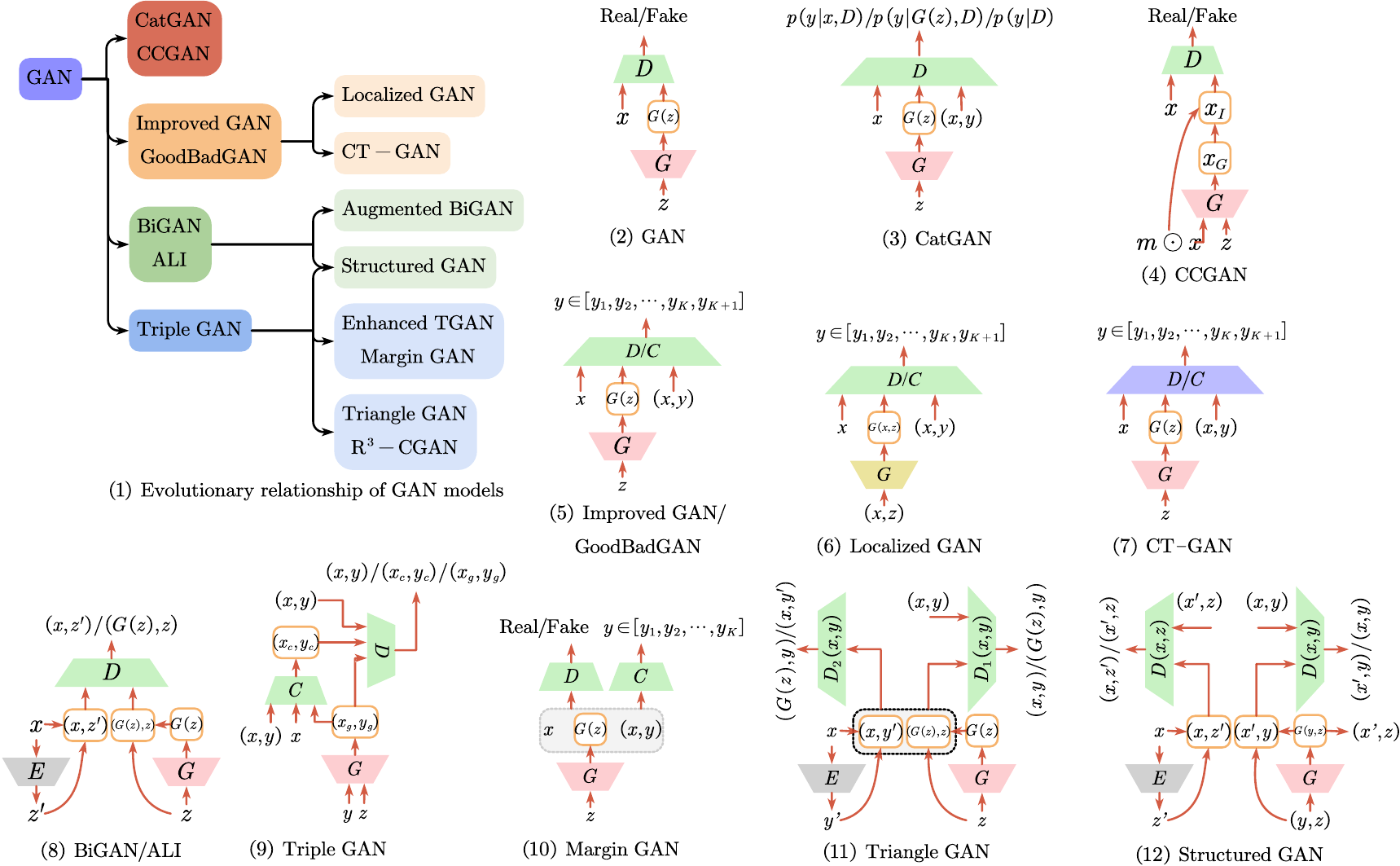}
		\caption{A glimpse of the diverse range of architectures used for GAN-based deep generative semi-supervised methods. The characters `$`D,G$" and  ``$E$" represent \emph{Discriminator}, \emph{Generator} and \emph{Encoder}, respectively. In Figure (6), Localized GAN is equipped with a local generator $G(x,z)$, so we use the yellow box to distinguish it. Similarly, in CT-GAN, the purple box is used to denote a discriminator that introduces consistency constraint.
		}
		\label{fig:generativeModel}
	\end{figure*}
	
	\subsection{Semi-supervised GANs}\label{sec:semiGAN}
	A typical GAN \cite{DBLP:conf/nips/GoodfellowPMXWOCB14} consists of a generator $G$ and a discriminator $D$ (see Fig.~\ref{fig:generativeModel}(2)). The goal of $G$ is to learn a distribution $p_g$ over data $x$ given a prior on input noise variables $p_z(z)$. The fake samples $G(z)$ generated by the generator $G$ are used to confuse the discriminator $D$. The discriminator $D$ is used to maximize the distinction between real training samples $x$ and fake samples $G(z)$. As we can see, $D$ and $G$ play the following two-player minimax game with the value function $V(G,D)$:
	\begin{align}
		\min\limits_G \max\limits_D V(D, G) = \mathbb{E}_{x \sim p(x)}[\text{log}D(x)] \nonumber \\
		+ \mathbb{E}_{z \sim p_{z}}[\text{log}(1 - D(G(z)))].
		\label{equ:gan}
	\end{align}
	
	Since GANs can learn the distribution of real data from unlabeled samples, it can be used to facilitate SSL. There are many ways to use GANs in SSL settings. Inspired by \cite{Schoneveld2017SemiSupervisedLW}, we have identified four main themes in how to use GANs for SSL, \ie, (a) re-using  the features from the discriminator, (b) using GAN-generated samples to regularize a classifier,  (c) learning an inference model, and (d) using samples produced by a GAN as additional training data.

	A simple SSL approach is to combine supervised and unsupervised loss during training. \cite{DBLP:journals/corr/RadfordMC15} demonstrates that a
	hierarchical representation of the GAN-discriminator is useful for object classification. These findings indicate that a simple and efficient SSL method can be provided by combining an unsupervised GAN value function with a supervised classification objective function, \eg, $\mathbb{E}_{(x,y)\in X_l}[\log D(y|x)]$. In the following, we review several representative methods of  semi-supervised GANs.

	\textbf{CatGAN.}
	Categorical Generative Adversarial Network (CatGAN) \cite{DBLP:journals/corr/Springenberg15} modifies the GAN's objective function to take into account the mutual information between observed examples and their predicted categorical class distributions. In particular, the optimization problem is different from the standard GAN (Eq.~(\ref{equ:gan})). The structure is illustrated in Fig.~\ref{fig:generativeModel}(3). This method aims to learn a discriminator which distinguishes the samples into $K$ categories by labeling $y$ to each $x$, instead of learning a binary discriminator value function. Moreover, in the CatGAN discriminator loss function, the supervised loss is also a cross-entropy term between the predicted conditional distribution $p(y|x,D)$ and the true label distribution of examples. It consists of three parts: (1) entropy $H[p(y|x,D)]$ which to obtain certain category assignment for samples; (2) $H[p(y|G(z), D)]$ for uncertain predictions from generated samples; and (3) the marginal class entropy $H[p(y|D)]$ to uniform usage of all classes. The proposed framework uses the feature space learned by the discriminator for the final learning task.  For the labeled data, the supervised loss is also a cross-entropy term between the conditional distribution $p(y|x,D)$ and the samples' true label distribution.

	\textbf{CCGAN.}
	Context-Conditional Generative Adversarial Networks (CCGAN) \cite{DBLP:journals/corr/DentonGF16} is proposed to use an adversarial loss for harnessing unlabeled image data based on image in-painting. The architecture of the CCGAN is shown in Fig.~\ref{fig:generativeModel}(4). The main highlight of this work is context information provided by the surrounding parts of the image. The method trains a GAN where the generator is to generate pixels within a missing hole. The discriminator is to discriminate between the real unlabeled images and these in-painted images. More formally, $m\odot x$ as input to a generator, where $m$ denotes a binary mask to drop out a specified portion of an image and $\odot$ denotes element-wise multiplication. Thus the in-painted image $x_I=(1-m)\odot x_G+m\odot x$ with generator outputs $x_G=G(m\odot x, z)$. The in-painted examples provided by the generator cause the discriminator to learn features that generalize to the related task of classifying objects. The penultimate layer of components of the discriminator is then shared with the classifier, whose cross-entropy loss is used combined with the discriminator loss.

	\textbf{Improved GAN.}
	There are several methods to adapt GANs into a semi-supervised classification scenario. CatGAN \cite{DBLP:journals/corr/Springenberg15} forces the discriminator to maximize the mutual information between examples and their predicted class distributions instead of training the discriminator to learn a binary classification. To overcome the learned representations' bottleneck of CatGAN, Semi-supervised GAN (SGAN) \cite{DBLP:journals/corr/Odena16a} learns a generator and a classifier simultaneously. The classifier network can have $(K+1)$ output units corresponding to $[y_1,y_2,\ldots, y_K, y_{K+1}]$, where the $y _{K+1}$ represents the outputs  generated by $G$. Similar to SGAN, Improved GAN \cite{DBLP:conf/nips/SalimansGZCRCC16} solves a $(K+1)$-class classification problem. The structure of Improved GAN is shown in Fig.~\ref{fig:generativeModel}(5). Real examples for one of the first $K$ classes and the additional $(K+1)$th class consisted of the synthetic images generated by the generator $G$. This work proposes the improved techniques to train the GANs, \ie, feature matching, minibatch discrimination, historical averaging one-sided label smoothing, and virtual batch normalization, where feature matching is used to train the generator. It is trained by minimizing the discrepancy between features of the real and the generated examples, that is $\|\mathbb{E}_{x\in X} D(x)-\mathbb{E}_{z\sim p(z)}D(G(z))\|_2^2$, rather than maximizing the likelihood of its generated examples classified to  $K$ real classes. The loss function for training the classifier becomes
	\begin{equation}
		\begin{aligned}
			&\max_D \mathbb{E}_{(x,y)\sim p(x,y)}\log p_D(y|x,y\le K) \\
			&+\mathbb{E}_{x\sim p(x)}\log p_D(y\le K|x)+\mathbb{E}_{x\sim p_G}\log p_D(y=K+1|x),	
		\end{aligned}
		\label{equ:improvedGAN}
	\end{equation}
	where the first term of Eq.~(\ref{equ:improvedGAN}) denotes the supervised cross-entropy loss, The last two terms of Eq.~(\ref{equ:improvedGAN}) are the unsupervised losses from the unlabeled and generated data, respectively.
	
	\textbf{GoodBadGAN.}
	GoodBadGAN \cite{DBLP:conf/nips/DaiYYCS17} realizes that the generator and discriminator in \cite{DBLP:conf/nips/SalimansGZCRCC16} may not be optimal simultaneously, \ie, the discriminator achieves good performance in SSL, while the generator may generate visually unrealistic samples. The structure of GoodBadGAN is shown in Fig.~\ref{fig:generativeModel}(5).
	The method gives theoretical justifications of why using bad samples from the generator can boost the SSL performance. Generally, the generated samples, along with the loss function (Eq.~(\ref{equ:improvedGAN})), can force the boundary of the discriminator to lie between the data manifolds of different categories, which improves the generalization of the discriminator. Due to the analysis, GoodBadGAN learns a bad generator by explicitly adding a penalty term $\mathbb{E}_{x\sim p_G}\log p(x)\mathbb{I}[p(x)>\epsilon]$ to generate bad samples, where $\mathbb{I}[\cdot]$ is an indicator function and $\epsilon$ is a threshold, which ensures that only high-density samples are penalized while low-density samples are unaffected. Further, to guarantee the strong true-fake belief in the optimal conditions, a conditional entropy term $\mathbb{E}_{x \sim p_x} \sum_{k=1}^{K} \log p_D(k|x)$ is added to the discriminator objective function in  Eq.~(\ref{equ:improvedGAN}).
	
	\textbf{Localized GAN.}
	Localized GAN \cite{DBLP:conf/cvpr/QiZHEWH18} focuses on using local coordinate charts to parameterize local geometry of data transformations across different locations manifold rather than the global one. This work suggests that Localized GAN can help train a locally consistent classifier by exploring the manifold geometry. The architecture of Localized GAN is shown in Fig.~\ref{fig:generativeModel}(6). Like the methods  introduced in \cite{DBLP:conf/nips/SalimansGZCRCC16,DBLP:conf/nips/DaiYYCS17}, Localized GAN attempts to solve the $K+1$ classification problem that outputs the probability of $x$ being assigned to a class. For a classifier, $\nabla_z D(G(x,z))$ depicts the change of the classification decision on the manifold formed by $G (x,z)$, and the evolution of $D(\cdot)$ restricted on $G(x,z)$ can be written as
	\begin{equation}
		|D(G(x,z+\sigma z))-D(G(x,z))|^2\approx \|\nabla_x^G D(x)\|^2 \sigma z,
	\end{equation}
	which shows that penalizing $\|\nabla_x^G D(x)\|^2$ can train a robust classifier with a small perturbation $\sigma z$ on a manifold. This probabilistic classifier can be introduced  by adding $\sum_{k=1}^{K}\mathbb{E}_{x\sim p_x}\|\nabla_x^G \log p(y=k|x)\|^2$, where $\nabla_x^G \log p(y=k|x)$ is the gradient of the log-likelihood along with the manifold $G(x,z)$.
	
	\textbf{CT-GAN.}
	CT-GAN \cite{DBLP:conf/iclr/WeiGL0W18} combines consistency training with WGAN  \cite{DBLP:journals/corr/ArjovskyCB17} applied to semi-supervised classification problems. And the structure of CT-GAN is shown in Fig.~\ref{fig:generativeModel}(7). Following \cite{DBLP:conf/nips/GulrajaniAADC17}, this method also lays the Lipschitz continuity condition over the manifold of the real data to improve the improved training of WGAN. Moreover, CT-GAN devises a regularization over a pair of samples drawn near the manifold following the most basic definition of the 1-Lipschitz continuity. In particular, each real instance $x$ is perturbed twice and uses a Lipschitz constant to bound the difference between the discriminator's responses to the perturbed instances $x',x''$. Formally, since the value function of WGAN is
	\begin{equation}
		\min_G\max_D \mathbb{E}_{x\sim p_x}D(x)-\mathbb{E}_{z\sim p_z}D(G(z)),
		\label{equ:wgan}
	\end{equation}
	where $D$ is one of the sets of 1-Lipschitz function.
	The objective function for updating the discriminator include: (a) basic Wasserstein distance in Eq.~(\ref{equ:wgan}), (b) gradient penalty $GP|_{\hat{x}}$ \cite{DBLP:conf/nips/GulrajaniAADC17} used in the improved training of WGAN, where $\hat{x}=\epsilon x+(1-\epsilon)G(z)$, and (c) A consistency regularization $CT|_{x',x''}$. For semi-supervised classification, CT-GAN uses the Eq.~(\ref{equ:improvedGAN}) for training the discriminator instead of the Eq.(~\ref{equ:wgan}), and then adds the consistency regularization $CT|_{x',x''}$.

	\textbf{BiGAN.}
	Bidirectional Generative Adversarial Networks (BiGANs) \cite{DBLP:conf/iclr/DonahueKD17} is an unsupervised feature learning framework. The architecture of BiGAN is shown in Fig.~\ref{fig:generativeModel}(8). In contrast  to the standard GAN framework, BiGAN adds an encoder $E$ to this framework, which maps data $x$ to $z'$, resulting in a data pair $(x, z')$. The data pair $(x,z')$ and the data pair generated by generator $G$ constitute two kinds of true and fake data pairs. The BiGAN discriminator $D$ is to distinguish the true and fake data pairs. In this work, the value function for training the discriminator  becomes,
	\begin{equation}
		\begin{aligned}
			\min_{G,E}\max_{D} V(D,E,G)=\mathbb{E}_{x\sim \mathcal{X}}\underset{\log D(x,E(x))}{\underbrace{\left[ \mathbb{E}_{z\sim p_E\left( \cdot |x \right)}\left[ \log D\left( x,z \right) \right] \right] }}
			\\
			+\mathbb{E}_{z\sim p\left( z \right)}\underset{\log \left( 1-D\left( G\left( z \right) ,z \right) \right)}{\underbrace{\left[ \mathbb{E}_{x\sim p_G\left( \cdot |z \right)}\left[ \log \left( 1-D\left( x,z \right) \right) \right] \right] }}.
		\end{aligned}
		\label{equ:bigan}
	\end{equation}

	\textbf{ALI.}
	Adversarially Learned Inference (ALI) \cite{DBLP:conf/iclr/DumoulinBPLAMC17} is a GAN-like adversarial framework based on the combination of an inference network and a generative model.
	This framework consists of three networks, a generator, an inference network and a discriminator.
	The generative network $G$ is used as a decoder to map latent variables $z$ (with a prior distribution) to data distribution $x'=G(z)$, which can  be  formed as joint pairs $(x',z)$. The inference network $E$ attempts to encode training samples $x$ to latent variables $z'=E(x)$, and joint pairs $(x,z')$ are similarly drawn. The discriminator network $D$ is required to distinguish the joint pairs $(x,z')$ from the joint pairs $(x',z)$. As discussed above, the central  architecture of ALI is regarded as similar to the BiGAN's (see Fig.~\ref{fig:generativeModel}(8). In semi-supervised settings, this framework adapts the discriminator network proposed in \cite{DBLP:conf/nips/SalimansGZCRCC16}, and shows promising performance on semi-supervised benchmarks on SVHN and CIFAR 10. The objective function can be rewritten as an extended version similar to Eq.~(\ref{equ:bigan}).
	
	\textbf{Augmented BiGAN.}
	Kumar \etal \cite{DBLP:conf/nips/KumarSF17} propose an extension of BiGAN called  Augmented BiGAN for SSL. This framework has a BiGAN-like architecture that consists of an encoder, a  generator and a discriminator. Since trained GANs produce realistic images, the generator can be considered to obtain the tangent spaces of the image manifold. The estimated tangents infer the desirable invariances which can be injected into the discriminator to improve SSL performance. In particular, the Augmented BiGAN uses feature matching loss \cite{DBLP:conf/nips/SalimansGZCRCC16}, $\|\mathbb{E}_{x\in  X}D(E(x),x)-\mathbb{E}_{z\sim p(z)}D(z,G(z))\|^2_2$ to optimize the generator network and the encoder network. Moreover, to avoid the issue of  class-switching (the class of $G(E(x))$ changed during the decoupled training),  a third pair $(E(x), G(E(x)))$ loss term $\mathbb{E}_{x\sim p( x)}[ \log ( 1-D( E( x) ,G_x( E( x) ) )) ] $ is added to the objective function Eq.~(\ref{equ:bigan}).
	
	\textbf{Triple GAN.}
	Triple GAN \cite{DBLP:conf/nips/LiXZZ17} is presented to address the issue that the generator and discriminator of GAN have incompatible loss functions, \ie, the generator and the discriminator can not be optimal at the same time \cite{DBLP:conf/nips/SalimansGZCRCC16}. The problem has been mentioned in \cite{DBLP:conf/nips/DaiYYCS17}, but the solution is different. As shown in Fig.~\ref{fig:generativeModel}(9), the Triple GAN tackles  this problem by playing a three-player game. This three-player framework consists of three parts, a generator $G$ using a conditional network to generate the corresponding fake samples for the true labels, a classifier $C$ that generates pseudo labels for given real data, and a discriminator $D$ distinguishing whether a data-label pair is from the real-label dataset or not. This Triple GAN loss function may be written as
	\begin{equation}
		\begin{aligned}
			\min_{C,G}\max_{D} V(C,G,D) =\mathbb{E}_{\left( x,y \right) \sim p\left( x,y \right)}\left[ \log D\left( x,y \right) \right]
			\\
			+ ~\alpha \mathbb{E}_{\left( x,y \right) \sim p_c\left( x,y \right)}\left[ \log \left( 1-D\left( x,y \right) \right) \right]
			\\
			+\left( 1-\alpha \right) \mathbb{E}_{\left( x,y \right) \sim p_g\left( x,y \right)}\left[ \log \left( 1-D\left( G\left( y,z \right) ,y \right) \right) \right],
		\end{aligned}
		\label{equ:tripleGAN}
	\end{equation}
	where $D$ obtains label information about unlabeled data from the classifier $C$ and forces the generator  $G$ to generate the realistic image-label samples.
	
	\textbf{Enhanced TGAN.}
	Based on the architecture of Triple GAN \cite{DBLP:conf/nips/GanCWPZLLC17}, Enhanced TGAN \cite{DBLP:conf/cvpr/WuDLL0W19} modifies the Triple-GAN by re-designing the generator loss function and the classifier network. The generator generates  images conditioned on class distribution and is regularized by class-wise mean feature matching. The classifier network includes two classifiers that collaboratively learn to provide more categorical information for generator training. Additionally, a semantic matching term is added to enhance the semantics consistency with respect to the generator and the classifier network. The discriminator $D$ learned to distinguish the labeled data pair $(x,y)$ from the synthesized data pair $(G(z),\tilde{y})$ and predicted data pair $(x,\bar{y})$. The corresponding objective  function is similar to Eq.~(\ref{equ:tripleGAN}), where $(G(z),\tilde{y})$ is sampling from the pre-specified distribution $p_g$, and $(x,\bar{y})$ denotes the predicted data pair determined by $p_c(x)$.
	
	\textbf{MarginGAN.}
	MarginGAN \cite{DBLP:conf/nips/DongL19} is another extension framework based on Triple GAN \cite{DBLP:conf/nips/GanCWPZLLC17}. From the perspective of classification margin, this framework works better than Triple GAN when used for semi-supervised classification.  The architecture of MarginGAN is presented in Fig.~\ref{fig:generativeModel}(10). MarginGAN includes three components like Triple GAN, a generator $G$ which tries to maximize the margin of generated samples, a classifier $C$ used for decreasing margin of fake images, and a discriminator $D$ trained as usual to distinguish real samples from fake images. This method solves the problem that performance is damaged due to the inaccurate pseudo label in SSL, and improves the accuracy rate of SSL.
	
	\textbf{Triangle GAN.}
	Triangle Generative Adversarial Network ($\triangle$-GAN) \cite{DBLP:conf/nips/GanCWPZLLC17} introduces a new architecture to match cross-domain  joint distributions. The architecture of the  $\triangle$-GAN is shown in Fig.~\ref{fig:generativeModel}(11). The $\triangle$-GAN can be considered as an extended version of BiGAN \cite{DBLP:conf/iclr/DonahueKD17} or ALI \cite{DBLP:conf/iclr/DumoulinBPLAMC17}. This framework is a four-branch model that consists of two generators $E$ and $G$, and two discriminators $D_1$ and  $D_2$. The two generators can learn two different  joint distributions by two-way matching between two different domains.  At the same time, the discriminators are used as an implicit ternary function, where $D_2$ determines whether the data pair is from $(x,y')$ or from $(G(z),y)$ , and $D_1$ distinguishes real data pair $(x,y)$ from the fake data pair $(G(z),y)$.
	
	\textbf{Structured GAN.}
	Structured GAN \cite{DBLP:conf/nips/DengZLYXZX17} studies the problem of semi-supervised conditional generative modeling based on designated semantics or structures. The architecture of Structured GAN (see  Fig.~\ref{fig:generativeModel}(12))  is similar to Triangle GAN \cite{DBLP:conf/nips/GanCWPZLLC17}. Specifically, Structured GAN assumes that the samples $x$ are generated conditioned on two independent latent variables, \ie, $y$ that encodes the designated semantics and $z$ contains other variation factors. Training Structured GAN involves solving two adversarial games that have their equilibrium concentrating at the true joint data distributions  $p(x,z)$ and $p(x,y)$. The synthesized data pair $(x',y)$ and $(x',z)$ are generated by generator $G(y,z)$, where $(x ',y)$ mix real sample pair $(x,y)$ together as input for training discriminator $D(x,y)$ and $(x'z)$ blends the $E$'s outputs pair $(x,z')$ for discriminator $D(x,z)$.
	
	\textbf{$\bm{R^3}$-CGAN.}
	Liu \etal  \cite{LiuYR3GAN20} propose a Class-conditional GAN with Random Regional Replacement (R3-regularization) technique, called $R^3$-CGAN. Their framework and training strategy relies on Triangle GAN \cite{DBLP:conf/nips/GanCWPZLLC17}. The $R^3$-CGAN architecture comprises four parts, a generator $G$ to synthesize fake images with specified class labels, a classifier $C$ to generate instance-label pairs of real unlabeled images with pseudo-labels, a discriminator $D_1$ to identify real or fake pairs,
	and another discriminator $D_2$ to distinguish two types of fake data.
	Specifically, CutMix \cite{DBLP:conf/iccv/YunHCOYC19}, a Random Regional Replacement strategy, is used to construct two types of between-class instances (cross-category instances and real-fake instances). These instances are used to regularize the classifier $C$ and discriminator $D_1$. Through the minimax game among the four players, the class-specific information is effectively used for the downstream.
	
	\textbf{Summary.}
	Comparing with the above discussed Semi-GANs methods, we find that the main difference lies in the number and type of the basic modules, such as generator, encoder, discriminator, and classifier. As shown in Fig.~\ref{fig:generativeModel}(1), we find the evolutionary relationship of the Semi-GANs models. Overall, CatGAN \cite{DBLP:journals/corr/Springenberg15} and CCGAN \cite{DBLP:journals/corr/DentonGF16} extend the basic GAN by including additional information in the model, such as category information and in-painted images. Based on Improved GAN \cite{DBLP:conf/nips/SalimansGZCRCC16}, Localized GAN \cite{DBLP:conf/cvpr/QiZHEWH18} and CT-GAN \cite{DBLP:conf/iclr/WeiGL0W18} consider the local information and consistency regularization, respectively. BiGAN \cite{DBLP:conf/iclr/DonahueKD17}  and ALI \cite{DBLP:conf/iclr/DumoulinBPLAMC17} learn an inference model during the training process by adding an Encoder module. In order to solve the problem that the generator and discriminator can not be optimal at the same time, Triple-GAN \cite{DBLP:conf/nips/LiXZZ17} adds an independent classifier instead of using a discriminator as a classifier.
	
	%fig: semi-supervised VAE models
	\subsection{Semi-supervised VAE}\label{sec:semiVAE}
	\begin{figure*}[!t]
		\centering
		\includegraphics[width=7.0in]{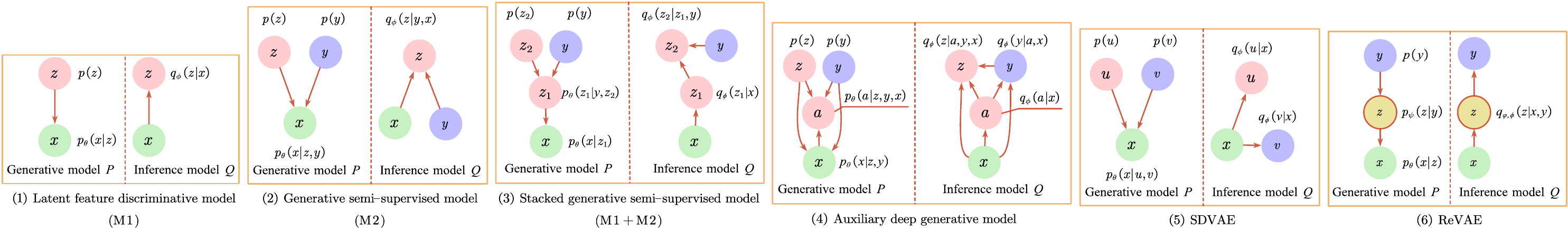}
		\caption{A glimpse of probabilistic graphical models used for VAE-based deep generative semi-supervised methods. Each method contains two models, the generative model $P$ and the inference model $Q$. The variational parameters $\theta$ and $\phi$ are learned jointly by the incoming connections (\ie, deep neural networks).}
		\label{fig:semiVAE}
	\end{figure*}
	
	Variational AutoEncoders (VAEs) \cite{DBLP:journals/corr/KingmaW13,DBLP:conf/icml/RezendeMW14} are flexible models which combine deep autoencoders with generative latent-variable models. The generative model captures representations of the distributions rather than the observations of the dataset, and defines the joint distribution in the form of $p(x,z)=p(z)p(x|z)$, where $p(z)$ is a prior distribution over latent variables $z$. Since the true posterior $p(z|x)$ is generally intractable, the generative model is trained with the aid of an approximate posterior distribution $q(z|x)$. The architecture of VAEs is a two-stage network, an encoder to construct a variational approximation $q(z|x)$ to the posterior $p(z|x)$, and a decoder to parameterize the likelihood $p(x|z)$. The variational approximation of the posterior maximizes the marginal likelihood, and the evidence lower bound (ELBO) may be written as
	\begin{align}
		\log p(x)=\log \mathbb{E}_{q(z|x)}[\frac{p(z)p(x|z)}{q(z|x)}]\\
		\geq \mathbb{E}_{q(z|x)}[\log p(z)p(x|z)-\log q(z|x)].
	\end{align}

	There are three reasons why latent-variable models can be useful for SSL: (a) It is a natural way to incorporate unlabeled data, (b) The ability to disentangle representations can be easily implemented via the configuration of latent variables, and (c) It also allows us to use variational neural methods. In the following, we review several representative latent variable methods for semi-supervised learning.

	\textbf{SSVAEs.}
	SSVAEs denotes the VAE-based generative models with latent encoder representation proposed in \cite{DBLP:conf/nips/KingmaMRW14}. 
	The first one, i.e., the latent-feature discriminative model, referred to M1 \cite{DBLP:conf/nips/KingmaMRW14}, can provide more robust latent features with a deep generative model of the data. As shown in Fig.~\ref{fig:semiVAE}(1), $p_{\theta}(x|z)$ is a non-linear transformation, \eg, a deep neural network. Latent variables $z$ can be selected as a  Gaussian distribution or a Bernoulli distribution. An approximate sample of the posterior distribution  on the latent variable $q_{\phi}(z|x)$ is used as the classifier feature for the class label $y$. The second one, namely Generative semi-supervised model, referred to M2 \cite{DBLP:conf/nips/KingmaMRW14}, describes the data generated by a latent class variable $y$ and a continuous latent variable $z$, is expressed as $p_{\theta}(x|z,y)p(z)p(y)$ (as depicted in Fig.~\ref{fig:semiVAE}(2)). $p(y)$ is the  multinomial distribution, where the class labels $y$ are treated as latent variables for unlabeled data. $p_{\theta}(x|z,y)$ is a suitable likelihood function. The inferred posterior distribution $q_{\phi}(z|y,x)$ can predict any missing labels.
	
	Stacked generative semi-supervised model, called M1+M2, uses the  generative model M1 to learn the new latent representation $z_1$,  and uses the  embedding from $z_1$ instead of the raw data $x$ to learn a generative semi-supervised model M2.  As shown in Fig.~\ref{fig:semiVAE}(3), the whole process can be abstracted as follows:
	\begin{equation}
		p _{\theta} (x,y,z_1,z_2)=p(y)p(z_2)p_{\theta}(z_1|y,z_2)p_{\theta}(x|z_1),
	\end{equation}
	where $p_{\theta}(z_1|y,z_2)$ and $p_{\theta}(x|z_1)$ are parameterised as deep neural networks. In all the above models, $q_{\phi}({z|x})$ is used to approximate the true posterior distribution $p({z|x})$, and following the variational principle, the boundary approximation lower bound of the model is derived to ensure that the approximate posterior probability is as close to the true posterior probability as possible.
	
	\textbf{ADGM.}
	Auxiliary Deep Generative Models (ADGM) \cite{DBLP:conf/icml/MaaloeSSW16} extends SSVAEs \cite{DBLP:conf/nips/KingmaMRW14} with auxiliary variables, as depicted in Fig.~\ref{fig:semiVAE}(4). The auxiliary variables  can improve the variational approximation and make the variational distribution more expressive by training deep generative models with multiple stochastic layers.
	
	Adding the auxiliary variable $a$ leaves the generative model of  $x,y$  unchanged while significantly improving the representative power of the posterior approximation. An additional  inference network is introduced such that:
	\begin{equation}
		q_{\phi}(a,y,z|x)=q_{\phi}(z|a,y,x)q_{\phi}(y|a,x)q_{\phi}(a|x).
	\end{equation}
	The framework has the generative model $p$ defined as $p_{\theta}(a)p_{\theta}(y)p_{\theta}(z)p_{\theta}(x|z,y)$, where $a,y,z$ are the auxiliary variable, class label, and latent features, respectively. Learning the posterior distribution is intractable. Thus we define the approximation as $q_{\phi}(a|{x})q_{\phi}({z}|y,{x})$ and a classifier $q_{\phi}(y|a,{x})$.
	The auxiliary unit $a$ actually introduces a class-specific latent distribution between $x$ and $y$, resulting in a more expressive distribution $q_{\phi}(y|a,x)$.  Formally, \cite{DBLP:conf/icml/MaaloeSSW16} employs the similar variational lower bound $\mathbb{E}_{q_{\phi}(a,z|x)}[\log p_{\theta}(a,x,y,z)-\log q_{\phi}(a,z|x,y)]$ on the marginal likelihood, with $q_{\phi}(a,z|x,y)=q_{\phi}(a|x)q_{\phi}(z|y,x)$. Similarly, the unlabeled ELBO is
	$ \mathbb{E}_{q_{\phi}(a,y,z|x)}[\log p_{\theta}(a,x,y,z)-\log q_{\phi}(a,y,z|x)]$
	with $q_{\phi}(a,y,z|x)=q_{\phi}(z|y,x)q_{\phi}(y|a,x)q_{\phi}(a|x)$.
	
	Interestingly, by reversing the direction of the dependence between $x$ and $a$, a model  similar to the stacked version of M1 and M2 is recovered (Fig.~\ref{fig:semiVAE}(3)), with what the authors  denote skip connections from the second stochastic layer and the labels to the inputs $x$. In  this case the generative model is affected, and the authors call this the Skip Deep Generative  Model (SDGM). This model is able to be trained end to end using stochastic gradient
	descent (SGD) (according to the  \cite{DBLP:conf/icml/MaaloeSSW16} the skip connection between $z$ and $x$ is crucial for training to converge).  Unsurprisingly, joint training for the model improves significantly upon the performance  presented in \cite{DBLP:conf/nips/KingmaMRW14}.

	\textbf{Infinite VAE.}
	Infinite VAE \cite{DBLP:conf/cvpr/AbbasnejadDH17} proposes a mixture model for combining variational autoencoders, a non-parametric Bayesian approach. This model can adapt to suit the input data by mixing coefficients by a Dirichlet process.It combines Gibbs sampling and variational inference that enables the model to learn the input's underlying structures efficiently.  Formally, Infinite VAE employs the mixing coefficients to assist SSL by combining the unsupervised generative model and a supervised discriminative model.
	The infinite mixture generative model as,
	\begin{equation}
		p(c,\pi,x,z)=p(c|\pi)p_{\alpha}(\pi)p_{\theta}(x|c,z)p(z),
	\end{equation}
	where $c$ denotes the assignment matrix for each instance to a VAE component where the VAE-$i$ can  best reconstruct instance $i$. $\pi$ is the mixing coefficient prior for $c$,  drawn from a Dirichlet distribution with parameter $\alpha$. Each latent variable $z_i$ in each VAE is drawn from a Gaussian distribution.
	
	\textbf{Disentangled VAE.}
	Disentangled VAE \cite{DBLP:conf/nips/NarayanaswamyPM17} attempts to learn disentangled representations using partially-specified graphical model structures and distinct encoding aspects of the data into separate variables. It explores  the graphical model for modeling a general dependency on observed and unobserved latent variables with neural networks, and a stochastic computation graph \cite{DBLP:conf/nips/SchulmanHWA15} is used to infer with and train the resultant generative model. For this purpose, importance sampling estimates are used to maximize the lower bound of both the supervised and semi-supervised likelihoods. Formally, this framework considers the conditional probability $q_{y,z|x}$, which has a factorization $q_{\phi}(y,z|x)=q_{\phi}(y|x,z)q_{\phi}(z|x)$ rather than  $q_{\phi}(y,z|x)=q_{\phi}(z|x,y)q_{\phi}(y|x)$ in  \cite{DBLP:conf/nips/KingmaMRW14},  which means we can no longer compute a simple Monte Carlo estimator by sampling from the unconditional distribution $q_{\phi}(z|x)$. Thus, the variational lower bound for supervised term expand below,
	\begin{equation}
		\mathbb{E}_{q_{\phi}(z|x,y)}[\log p_{\theta}(x|y,z)p(y)p(z)-q_{\phi}(y,z|x)].
	\end{equation}
	
	\textbf{SDVAE.}
	Semi-supervised Disentangled VAE (SDVAE) \cite{DBLP:journals/isci/LiPWPYC19} incorporates the label information to the latent representations by encoding the input disentangled representation and non-interpretable representation. The disentangled variable captures categorical information, and the non-interpretable variable consists of other uncertain information from the data. As shown in Fig.~\ref{fig:semiVAE}(5), SDVAE assumes the disentangled variable $v$ and the non-interpretable variable $u$ are independent condition on $x$, \ie,  $q_{\phi}(u,v|x)=q_{\phi}(u|x)q_{\phi}(v|x)$. This means that $q_{\phi}(v|x)$ is the encoder for the disentangled representation, and  $q_{\phi}(u|x)$ denotes the encoder for non-interpretable representation. Based on those assumptions, the variational lower bound is written as:
	\begin{equation}
		\mathbb{E}_{q(u|x),q(v|x)}[\log p(x|u,v)p(v)p(u)-\log q(u|x)q(v|x)].
	\end{equation}

	%fig: consistency regularization models
	\begin{figure*}[!t]
		\centering
		\includegraphics[width=7.0in]{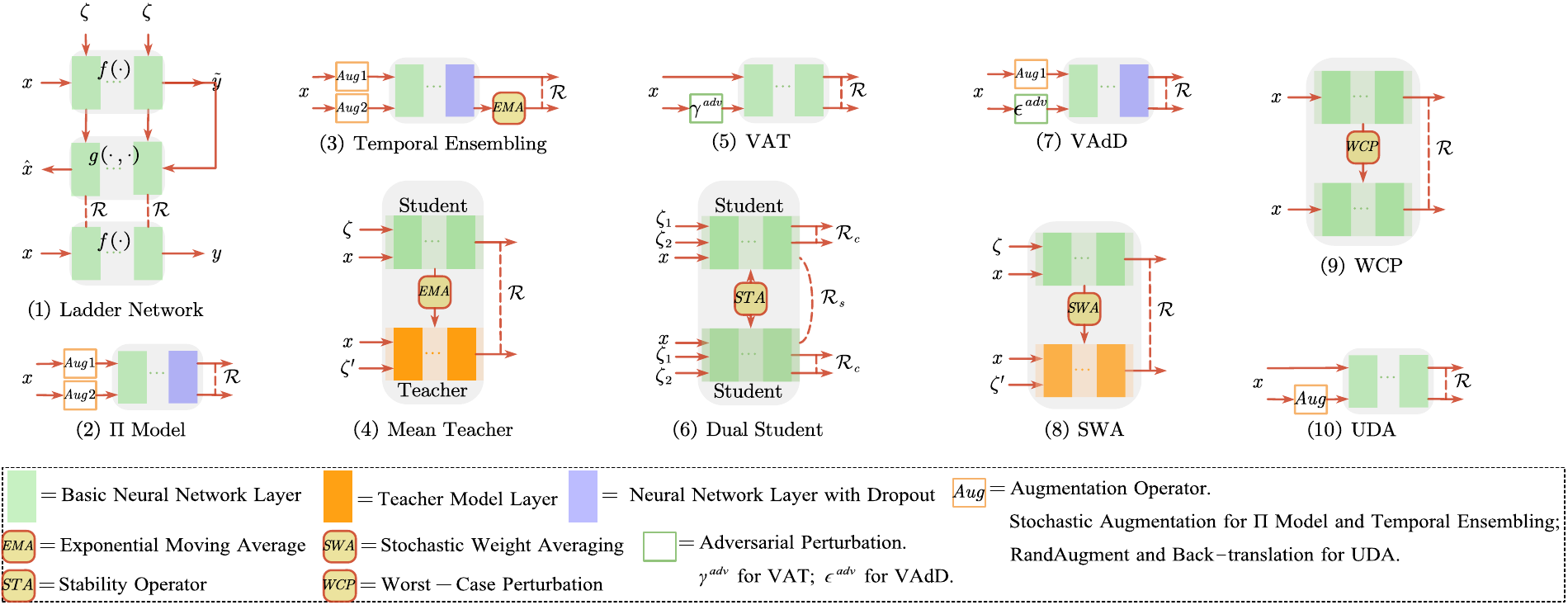}
		\caption{A glimpse of the diverse range of architectures used for consistency regularization semi-supervised methods. In addition to the identifiers in the figure, $\zeta$ denotes the perturbation noise, and $\mathcal{R}$ is the consistency constraint. }
		\label{fig:consistencyRegularization}
	\end{figure*}
	
	\textbf{ReVAE.}
	Reparameterized VAE (ReVAE) \cite{DBLP:journals/corr/abs-2006-10102} develops a novel way to encoder supervised information, which can capture label information through auxiliary variables instead of latent variables in the prior work  \cite{DBLP:conf/nips/KingmaMRW14}. The graphical model is illustrated in Fig.~\ref{fig:semiVAE}(6). In contrast to SSVAEs, ReVAE captures meaningful representations of data  with a principled variational objective. Moreover, ReVAE carefully designed the mappings between auxiliary and latent variables. In this model, a conditional generative model $p_{\psi}(z|y)$ is introduced to address the requirement for inference at test time. Similar to \cite{DBLP:conf/nips/KingmaMRW14} and  \cite{DBLP:conf/icml/MaaloeSSW16}, ReVAE treats $y$ as a known observation when the label is available in the supervised setting, and as an additional variable in the unsupervised case. In particular, the latent space can be partitioned into two disjoint subsets under the assumption that label information captures only specific aspects.
	
	\textbf{Summary.} As the name indicates, Semi-supervised VAE applies the VAE architecture for handling SSL problems. An advantage of these methods  is that  meaningful representations of data can be learned by the generative latent-variable models. The basic framework of these  Semi-supervised VAE methods is M2 \cite{DBLP:conf/nips/KingmaMRW14}. On the basis of the M2 framework, ADGM \cite{DBLP:conf/icml/MaaloeSSW16} and ReVAE \cite{DBLP:journals/corr/abs-2006-10102} consider introducing additional auxiliary variables, although the roles of the auxiliary variables in the two models are different. Infinite VAE \cite{DBLP:conf/cvpr/AbbasnejadDH17} is a hybrid of several VAE models to improve the performance of the entire framework. Disentangled VAE \cite{DBLP:conf/nips/NarayanaswamyPM17} and SDVAE \cite{DBLP:journals/isci/LiPWPYC19} solve the semi-supervised VAE problem by different disentangled methods. Under semi-supervised conditions, when a large number of labels are unobserved, the key to this kind of method is how to deal with the latent variables and label information.

	\section{Consistency Regularization}\label{sec:regularization}
	In this section, we introduce the consistency regularization methods for semi-supervised deep learning. In these methods, a consistency regularization term is applied to the final loss function to specify the prior constraints assumed by researchers. Consistency regularization is based on the manifold assumption or the smoothness assumption,  and describes a category of methods that the realistic perturbations of the data points should not change the output of the model \cite{DBLP:conf/nips/OliverORCG18}. Consequently, consistency regularization can be regarded to find a smooth manifold on which the dataset lies by leveraging the unlabeled data \cite{DBLP:conf/nips/BelkinN01}.
	
	The most common structure of consistency regularization SSL methods is the Teacher-Student structure. As a student, the model learns as before, and as a teacher, the model generates targets simultaneously. Since the model itself generates targets, they may be incorrect and then used by themselves as students for learning. In essence, the consistency regularization methods suffer from confirmation bias \cite{DBLP:conf/nips/TarvainenV17}, a risk that can be mitigated by improving the target's quality.  Formally, following \cite{DBLP:conf/iccv/KeWYRL19}, we assume that  dataset $X$ consists of a labeled subset $X_l$ and an unlabeled subset $X_u$. Let $\theta '$ denote the weight of the target, and $\theta$ denote the weights of the basic student. The consistency constraint is defined as:
	\begin{equation}
		\mathbb{E}_{x\in X } \mathcal{R}(f(\theta, x), \mathcal{T}_x),
		\label{equ:consistency_cost}
	\end{equation}
	where $f(\theta,x)$ is the prediction from model $f(\theta)$ for input $x$. $\mathcal{T}_x$ is the consistency target of the teacher. $\mathcal{R}(\cdot, \cdot)$ measures the distance between two vectors and is usually set to Mean Squared Error (MSE) or KL-divergence. Different consistency regularization techniques vary in how they generate the target. There are several ways to boost the target $\mathcal{T}_x$ quality. One strategy is to select the perturbation rather than additive or multiplicative noise carefully. Another technique is to consider the teacher model carefully instead of replicating the student model.
	
	\textbf{Ladder Network.}
	Ladder Network \cite{DBLP:conf/nips/RasmusBHVR15,DBLP:conf/icml/PezeshkiFBCB16} is the first successful attempt towards using a Teacher-Student model that is inspired by a deep denoising AutoEncoder. The structure of the Ladder Network is shown in Fig.~\ref{fig:consistencyRegularization}(1). In Encoder, noise $\zeta$ is injected into all hidden layers as the corrupted feedforward path $x+\zeta \rightarrow \frac{\text{Encoder}}{f( \cdot)}\rightarrow \tilde{z}_1\rightarrow \tilde{z}_2\,\,$ and shares the mappings $f(\cdot)$ with the clean encoder feedforward path $x\rightarrow  \frac{\text{Encoder}}{f(\cdot)}\rightarrow z_1\rightarrow z_2\rightarrow y$. The decoder path $\tilde{z}_1\rightarrow \tilde{z}_2 \rightarrow \frac{\text{Decoder}}{g( \cdot ,\cdot )}\rightarrow \hat{z}_2\rightarrow \hat{z}_1
	$ consists of the denoising functions $g(\cdot, \cdot)$ and the unsupervised denoising square error $\mathcal{R}$ on each layer consider as consistency loss between $\hat{z}_i$ and $z_i$. Through latent skip connections, the ladder network is  differentiated  from regular denoising AutoEncoder. This feature allows the higher layer features to focus on more abstract invariant features for the task. Formally, the ladder network unsupervised training loss $\mathcal{L}_u$ or the consistency loss is computed as the MSE between the activation of the clean encoder $z_i$ and the reconstructed activations $\hat{z}_i$. Generally, $\mathcal{L}_u$ is
	\begin{equation}
		\mathbb{E}_{x\in X}\mathcal{R}\left( f\left( \theta ,x \right) ,g\left( f\left( \theta ,x+\zeta \right) \right) \right).
		\label{equ:ladderNetwork}
	\end{equation}

	\textbf{$\bm{\Pi}$ Model.}
	Unlike the perturbation used in Ladder Network, $\Pi$ Model  \cite{DBLP:conf/nips/SajjadiJT16} is to create two random augmentations of a sample for both labeled and unlabeled data. Some techniques with non-deterministic behavior, such as randomized data augmentation, dropout, and random max-pooling, make an input sample pass through the network several times, leading to different predictions. The structure of the $\Pi$ Model is shown in Fig.~\ref{fig:consistencyRegularization}(2).  In each epoch of the training process for $\Pi$ Model, the same unlabeled sample propagates forward twice, while random perturbations are introduced by data augmentations and dropout. The forward propagation of the same sample may result in different predictions, and the $\Pi$ Model expects the two predictions to be as consistent as possible. Therefore, it provides an unsupervised consistency loss function,
	\begin{equation}
		\mathbb{E}_{x\in X} \mathcal{R}(f(\theta,x,\zeta_1),f(\theta,x,\zeta_2)),
	\end{equation}
	which minimizes the difference between the two predictions.

	\textbf{Temporal Ensembling.}
	Temporal Ensembling \cite{DBLP:conf/iclr/LaineA17} is similar to the $\Pi$ Model, which forms a consensus prediction under different regularization and input augmentation conditions. The structure of Temporal Ensembling is shown in Fig.~\ref{fig:consistencyRegularization}(3). It modifies the $\Pi$ Model by leveraging the Exponential Moving Average (EMA) of past epochs predictions. In other words, while $\Pi$ Model needs to forward a sample twice in each iteration, Temporal Ensembling reduces this computational overhead by using EMA to accumulate the predictions over epochs as $\mathcal{T}_x$. Specifically, the ensemble outputs $Z_i$ is updated with the  network  outputs $z_i$ after each training epoch, \ie, $Z_i\gets \alpha Z_i+\left( 1-\alpha \right) z_i$, where $\alpha$ is a momentum term. During the training process, the $Z$ can be considered to contain an average ensemble of $f(\cdot)$ outputs due to Dropout  and stochastic augmentations. Thus, consistency loss is:
	\begin{equation}
		\mathbb{E}_{x\in X} \mathcal{R} (f(\theta, x, \zeta_1), \text{EMA}(f(\theta, x, \zeta_2))).
		\label{equ:temporalEnsembling}
	\end{equation}

	%Table 2 summary of consistency regularization models
	\begin{table*}
		\centering
		\caption{Summary of Consistency Regularization Methods}
		\label{tab:consistency}
		%\begin{tabular}{>{\centering\arraybackslash}m{1.9cm}|>{\centering\arraybackslash}m{7cm}|>{\centering\arraybackslash}m{2cm}|c}
		\begin{tabular}{c|c|c|c}
			\hline
			\textbf{Methods} & \textbf{Techniques} & \textbf{Transformations} & \textbf{Consistency Constraints} \\
			\hline
			\makecell[c]{Ladder \\ Network} &\makecell[c]{Additional Gaussian Noise \\ in every neural layer} & input &$\mathbb{E}_{x\in X}\mathcal{R}( f( \theta ,x ) ,f( \theta ,x+\zeta ) ) $ \\
			\hline
			$\Pi$ Model &Different Stochastic Augmentations  & input & $\mathbb{E}_{x\in X}\mathcal{R}\left( f\left( \theta ,x,\zeta _1 \right) ,f\left( \theta ,x,\zeta _2 \right) \right)$ \\
			\hline
			\makecell[c]{Temporal \\ Ensembling} & \makecell[c]{Different Stochastic Augmentation \\ and EMA the predictions} & \makecell[c]{input, \\ predictions}  &  $\mathbb{E}_{x\in X}\mathcal{R}\left( f\left( \theta ,x,\zeta _1 \right) ,\text{EMA}\left( f\left( \theta ,x,\zeta _2 \right) \right) \right) $\\
			\hline
			Mean Teacher &\makecell[c]{Different Stochastic Augmentation \\ and EMA the weights}  & \makecell[c]{input,\\ weights } & $\mathbb{E}_{x\in X}\mathcal{R}\left( f\left( \theta ,x,\zeta \right) ,f\left( \text{EMA}\left( \theta \right) ,x,\zeta \right) \right)
			$
			\\
			\hline
			VAT & Adversarial perturbation & input  & $\mathbb{E}_{x\in X}\mathcal{R}\left( f\left( \theta ,x \right) ,f\left( \theta ,x,\gamma ^{adv} \right) \right)
			$
			\\
			\hline
			Dual Student &\makecell[c]{Stable sample \\ and stabilization constraint} & \makecell[c]{input, \\ weights}  &  $\mathbb{E}_{x\in X}\mathcal{R}\left( f\left( \text{STA}\left( \theta ,x_i \right) ,\zeta _1 \right) ,f\left( \text{STA}\left( \theta ,x_j \right) ,\zeta _2 \right) \right)$
			\\
			\hline
			SWA & Stochastic Weight Averaging  & \makecell[c]{input,\\ weights}  & $\mathbb{E}_{x\in X}\mathcal{R}\left( f\left( \theta ,x \right) ,f\left( \text{SWA}\left( \theta \right) ,x,\zeta \right) \right) $
			\\
			\hline
			VAdD & \makecell[c]{Adversarial perturbation and Stochastic \\ Augmentation (dropout mask)} & \makecell[c]{input,\\ weights} &$\mathbb{E}_{x\in X}\mathcal{R}\left( f\left( \theta ,x,\epsilon ^s \right) ,f\left( \theta ,x,\epsilon ^{adv} \right) \right) $
			\\
			\hline
			UDA & \makecell[c]{AutoAugment/RandAugment for image; \\ Back-Translation for text} & input  & $
			\mathbb{E}_{x\in X}\mathcal{R}\left( f\left( \theta ,x \right) ,f\left( \theta ,x,\zeta \right) \right) $
			\\
			\hline
			WCP & \makecell[c]{Additive perturbation on network weights,  \\DropConnect perturbation for network structure} & \makecell[c]{input, \\ network structure} &$\mathbb{E}_{x\in X}\mathcal{R}\left( f\left( \theta ,x \right) ,g\left( \theta +\zeta ,x \right) \right) $\\
			\hline
		\end{tabular}
	\end{table*}
	
	\textbf{Mean Teacher.}
	Mean Teacher \cite{DBLP:conf/nips/TarvainenV17} averages model weights using EMA over training steps and tends to produce a more accurate model instead of  directly using output predictions. The structure of Mean Teacher is shown in Fig.~\ref{fig:consistencyRegularization}(4). Mean Teacher consists of two models called Student and Teacher. The student model is a regular model similar to the $\Pi$ Model, and the teacher model has the same architecture as the student model with exponential moving averaging the student weights. Then Mean Teacher applied a consistency constraint  between the two predictions of student and teacher:
	\begin{equation}
		\mathbb{E}_{x\in X}\mathcal{R}(f(\theta,x,\zeta),f(\text{EMA}(\theta),x,\zeta')).
		\label{equ:meanTeacher}
	\end{equation}

	\textbf{VAT.}
	Virtual Adversarial Training \cite{DBLP:journals/pami/MiyatoMKI19} proposes the concept of adversarial attack for consistency regularization.  The structure of VAT is shown in Fig.~\ref{fig:consistencyRegularization}(5). This technique aims to generate an adversarial transformation of a sample, which can change the model prediction. Specifically, the adversarial training technique is used to find the optimal adversarial perturbation $\gamma$ of a real input instance $x$ such that $\gamma \leq \delta$. Afterward, the consistency constraint is applied between the model's output of the original input sample  and  the perturbed one, \ie,
	\begin{equation}
		\mathbb{E}_{x\in X}\mathcal{R}( f( \theta ,x) ,g( \theta ,x+\gamma ^{adv} )),
		\label{equ:VAT}
	\end{equation}
	where $\gamma^{adv}=\argmax_{\gamma;\|\gamma\|\leqslant \delta}\mathcal{R}(f(\theta,x),g(\theta,x+\gamma))$.
	
	\textbf{Dual Student.}
	Dual Student \cite{DBLP:conf/iccv/KeWYRL19} extends the Mean Teacher model by replacing the teacher with another student. The structure of Dual Student is shown in Fig.~\ref{fig:consistencyRegularization}(6). The two students start from different initial states and are optimized through individual paths during training. The authors also defined a novel concept, ``stable sample", along with a stabilization constraint to avoid the performance bottleneck produced by a coupled EMA Teacher-Student model. Hence, their weights may not be tightly coupled, and each learns its own knowledge. Formally, Dual Student checks whether $x$ is a stable sample for student $i$:
	\begin{equation}
		\mathcal{C}_{x}^{i}=\left\{ p_{x}^{i}=p_{\bar{x}}^{i} \right\} _1\&\left( \left\{ \mathcal{M}_{x}^{i}>\xi \right\} _1\left\| \left\{ \mathcal{M}_{\bar{x}}^{i}>\xi \right\} _1 \right. \right),
	\end{equation}
	where $\mathcal{M}_{x}^{i}=\left\| f\left( \theta ^i,x \right) \right\| _{\infty}$, and the stabilization constraint:
	\begin{equation}
		\mathcal{L}^i_{sta}=\begin{cases}
			\left\{ \varepsilon ^i>\varepsilon ^j \right\} _1\mathcal{R}\left( f\left( \theta ^i,x \right) ,f\left( \theta ^j,x \right) \right) &  \mathcal{C}^i=\mathcal{C}^j=1\\
			\mathcal{C}^i\mathcal{R}\left( f\left( \theta ^i,x \right) ,f\left( \theta ^j,x \right) \right) & \text{otherwise}\\
		\end{cases}.
		\label{equ:dualStudent}
	\end{equation}

	\textbf{SWA.}
	Stochastic Weight Averaging (SWA) \cite{DBLP:conf/uai/IzmailovPGVW18} improves generalization than conventional training. The aim is to average multiple points along the trajectory of stochastic gradient descent (SGD) with a cyclical learning rate and seek much flatter solutions than SGD. The consistency-based SWA \cite{DBLP:conf/iclr/AthiwaratkunFIW19} observes that SGD fails to converge on the consistency loss but continues to explore many solutions with high distances apart in predictions on the test data. The structure of SWA is shown in Fig.~\ref{fig:consistencyRegularization}(7). The SWA procedure also approximates the Teacher-Student approach, such as  $\Pi$ Model and Mean Teacher with a single model. The authors propose fast-SWA, which adapts the SWA to increase the distance between the averaged weights by using a longer cyclical learning rate schedule and diversity of the corresponding predictions by averaging multiple network weights  within each cycle. Generally, the consistency loss can be rewritten as follows:
	\begin{equation}
		\mathbb{E}_{x\in X}\mathcal{R}(f(\theta, x), f(\text{SWA}(\theta),x,\zeta)).
		\label{equ:swa}
	\end{equation}

	%VAdD
	\textbf{VAdD.}
	In VAT, the adversarial perturbation is defined as an additive noise unit vector applied to the input or embedding spaces, which has improved the generalization performance of SSL. Similarly, Virtual  Adversarial Dropout (VAdD) \cite{DBLP:conf/aaai/ParkPSM18} also employs adversarial training in addition to the $\Pi$ Model.  The structure of VAdD  is shown in Fig.~\ref{fig:consistencyRegularization}(8). Following the design of $\Pi$ Model, the consistency constraint of VAdD  is computed from two different dropped networks: one dropped network uses a random dropout mask, and the other applies adversarial training to the optimized dropout network. Formally, $f(\theta, x, \epsilon)$ denotes an output of a neural network with a random dropout mask, and the consistency loss incorporated adversarial dropout  is described as:
	\begin{equation}
		\mathbb{E}_{x\in X}\mathcal{R}(f(\theta, x, \epsilon^s), f(\theta, x, \epsilon^{adv})),
		\label{equ:vadd}
	\end{equation}
	where $\epsilon^{adv}=\argmax_{\epsilon;\|\epsilon^s-\epsilon\|_2\le \delta H}\mathcal{R}(f(\theta,x,\epsilon^s),f(\theta,x,\epsilon))$; $f(\theta, x, \epsilon^{adv})$ represents an adversarial target; $\epsilon^{adv}$ is an adversarial dropout mask; $\epsilon^s$ is a sampled random dropout mask instance; $\delta$ is a hyperparameter controlling the intensity of the noise, and $H$ is the dropout layer dimension.
	
	\textbf{WCP.}
	A novel regularization mechanism for training deep SSL by minimizing the Worse-case Perturbation (WCP) is presented by Zhang \etal \cite{DBLP:conf/cvpr/ZhangLH20WCP}. The structure of WCP is shown in Fig.~\ref{fig:consistencyRegularization}(9). WCP considers two forms of WCP regularizations -- additive and DropConnect perturbations, which impose additive perturbation on network weights and make structural changes by dropping the network connections, respectively. Instead of generating an ensemble of randomly corrupted networks, the WCP suggests enhancing the most vulnerable part of a network by making the most resilient weights and connections against the worst-case perturbations.  It enforces an additive noise on the model parameters $\zeta$, along with a constraint on the norm of the noise. In this case, the WCP regularization becomes,
	\begin{equation}
		\mathbb{E}_{x\in X}\mathcal{R}(f(\theta, x), g(\theta+\zeta, x)).
	\end{equation}
	The second perturbation is at the network structure level by DropConnect, which drops some network connections. Specifically, for parameters $\theta$, the perturbed version is $(1-\alpha)\theta$, and the $\alpha=1$ denotes a dropped connection while $\alpha=0$ indicates an intact one. By applying the consistency constraint, we have
	\begin{equation}
		\mathbb{E}_{x\in X} \mathcal{R}(f(\theta,x), f((1-\alpha)\theta, x)).
	\end{equation}
	
	\textbf{UDA.}
	UDA stands for Unsupervised Data Augmentation \cite{DBLP:conf/nips/XieDHL020} for image classification and text classification. The structure of UDA is shown in Fig.~\ref{fig:consistencyRegularization}(10).
	This method investigates the role of noise injection in consistency training and substitutes simple noise operations with high-quality data augmentation methods, such as  AutoAugment \cite{DBLP:conf/cvpr/CubukZMVL19}, RandAugment \cite{DBLP:journals/corr/abs-1909-13719} for images, and Back-Translation \cite{DBLP:conf/acl/SennrichHB16,DBLP:conf/emnlp/EdunovOAG18} for text. Following the consistency regularization framework, the UDA \cite{DBLP:conf/nips/XieDHL020} extends the advancement in supervised data augmentation to SSL.
	As discussed above, let $f(\theta, x,\zeta)$ be the augmentation transformation from which one can draw an augmented example $(x,\zeta)$ based on an original example $x$. The consistency loss is:
	\begin{equation}
		\mathbb{E}_{x\in X} \mathcal{R}(f(\theta, x),f(\theta, x, \zeta)),
	\end{equation}
	where $\zeta$ represents the data augmentation operator to create an augmented version of an input $x$.
	
	\textbf{Summary.}
	The core idea of consistency regularization methods is that the output of the model remains unchanged under realistic perturbation. As shown in TABLE~\ref{tab:consistency}, Consistency constraints can be considered at three levels: input dataset, neural networks and training process. From the input dataset perspective, perturbations are usually added to  the input examples: additive noise, random augmentation, or even adversarial training. We can drop some layers or connections for the networks, as WCP \cite{DBLP:conf/cvpr/ZhangLH20WCP}. From the training process, we can use SWA to make the SGD fit the consistency training or EMA parameters of the model for some training epochs as new parameters.

	\section{Graph-based methods}\label{sec:graph}
	Graph-based semi-supervised learning (GSSL) has always been a hot subject for research with a vast number of successful models\cite{DBLP:conf/cvpr/IscenTAC19,DBLP:conf/cvpr/ChenMQXZ20,DBLP:conf/cvpr/LiLCCYY20} because of its wide variety of applicability. The basic assumption in GSSL is that a graph can be extracted from the raw dataset where each node represents a training sample, and each edge denotes some similarity measurement of the node pair. 
	%Most of these approaches can be divided into two main categories: graph regularization and graph embedding. The former uses Laplacian regularization specifically, assuming that connected nodes with a strongly connected edge share probably the same labels, such as label propagation (LP)~\cite{Zhu2002LearningFL}, Gaussian random fields (GRF)~\cite{zhu2003semi} and Local and Global Consistency (LGC)~\cite{DBLP:conf/nips/ZhouBLWS03}. The principal goal of the latter is to encode the nodes as small-scale vectors representing their role and the structure information of their neighborhood.
	In this section, we review graph embedding SSL methods, and the principal goal is to encode the nodes as small-scale vectors representing their role and the structure information of their neighborhood.
	
	%fig: graph-based models
	\begin{figure*}[!t]
		\centering
		\includegraphics[width=7.0in]{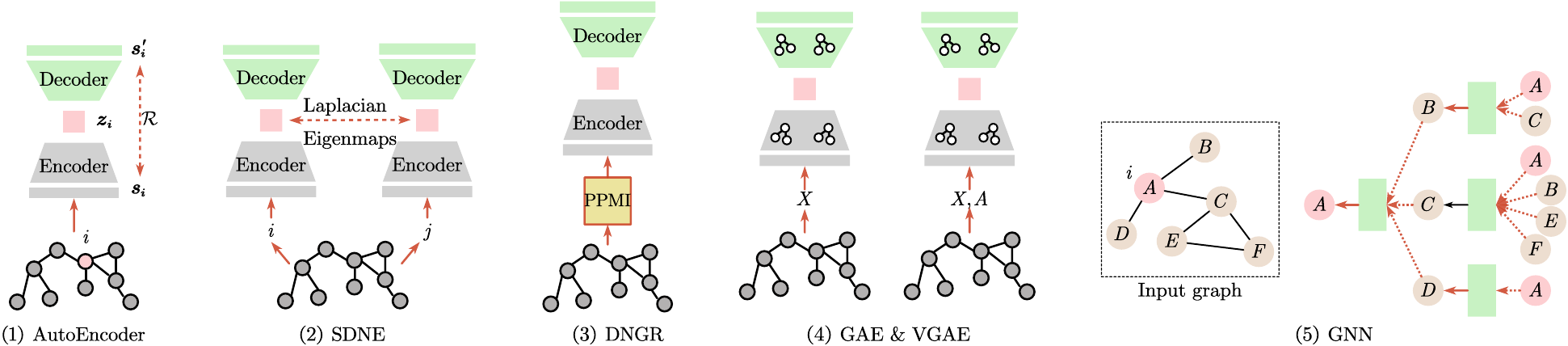}
		\caption{A glimpse of the diverse range of architectures used for graph-based semi-supervised methods. Specifically, in figure (3), ``PPMI" is short for positive pointwise mutual information. In figure (4), $A$ denotes the adjacency matrix, and in figure (5), pink A represents node A.}
		\label{fig:graph_based}
	\end{figure*}
	
	For graph embedding methods, we have the formal definition for the embedding on the node level. Given the graph $\mathcal{G}(\mathcal{V}, \mathcal{E})$, the node that has been embedded is a mapping result of $f_{\mathbf{z}}: v \rightarrow \mathbf{z}_v \in \mathbb{R}^{d}, \forall v \in {\mathcal{V}}$ such that $d \ll|{\mathcal{V}}|$, and the $f_{\mathbf{z}}$ function retains some of the measurement of proximity, defined in the graph $\mathcal{G}$.
	
	The unified form of the loss function is shown as Eq.~(\ref{general_graph_embedding}) for graph embedding methods.
	\begin{equation}
		\label{general_graph_embedding}
		\begin{aligned}
			\mathcal{L}(f_{\mathbf{z}}) =&\sum_{(x,y) \in X_l} (x,y,f_{\mathbf{z}})+\alpha \sum_{x\in X} \mathcal{R}(x,f_{\mathbf{z}}),
		\end{aligned}
	\end{equation}
	where $f_{\mathbf{z}}$ is the embedding function.
	
	Besides, we can divide graph embedding methods into shallow embedding and deep embedding based on the use or not use of deep learning techniques. The encoder's function, as a simple lookup function based on node ID, is built with shallow embedding methods including DeepWalk~\cite{DBLP:conf/kdd/PerozziAS14}, LINE~\cite{DBLP:conf/www/TangQWZYM15} and node2vec~\cite{DBLP:conf/kdd/GroverL16}, whereas the deep embedding encoder is far more complicated, and the deep learning frameworks have to make full use of node attributes.
	
	As deep learning progresses, recent GSSL research has switched from shallow embedding methods to deep embedding methods in which the $f_{\mathbf{z}}$ embedding term in the Eq.~(\ref{general_graph_embedding}) is focused on deep learning models. Two classes can be identified among all deep embedding methods: AutoEncoder-based methods and GNN-based methods.
	
	\subsection{AutoEncoder-based methods}
	AutoEncoder-based approaches also differ as a result of using a unary decoder rather than a pairwise method. More precisely, every node, ${i}$, is correlated to a neighborhood vector $\mathbf{s}_{i} \in \mathbb{R}^{|{\mathcal{V}}|}$. The $\mathbf{s}_{i}$ vector contains ${i}$'s similarity with all other graph nodes and acts as a high-dimensional vector representation of ${i}$ in the neighborhood.  The aim of the auto-encoding technique is to embed nodes using hidden embedding vectors like $\mathbf{s}_{i}$ in so as to reconstruct the original information from such embeddings (Fig.~\ref{fig:graph_based}(1)):
	
	\begin{equation}
		\label{autoencoder_embedding_goal}
		\operatorname{Dec}\left(\operatorname{Enc}\left(\mathbf{s}_{i}\right)\right)=\operatorname{Dec}\left(\mathbf{z}_{i}\right) \approx \mathbf{s}_{i}.
	\end{equation}
	In other words, the loss for these methods takes the following form:
	\begin{equation}
		\mathcal{L}=\sum_{{i} \in {V}}\left\|\operatorname{Dec}\left(\mathbf{z}_{i}\right)-\mathbf{s}_{i}\right\|_{2}^{2}.
	\end{equation}

	\textbf{SDNE.}
	Structural deep network embedding (SDNE) is developed by Wang ~\etal~\cite{wang2016structural} by using deep autoencoders to maintain the first and second-order network proximities. This is managed by optimizing the two proximities simultaneously. The process utilizes highly nonlinear functions to obtain its embedding. This framework consists of two parts: unsupervised and supervised. The first is an autoencoder to identify an embedding for a node to rebuild the neighborhood. The second is based on Laplacian Eigenmaps~\cite{belkin2002laplacian}, which imposes a penalty when related vertices are distant from each other.
	
	\textbf{DNGR.}
	(DNGR)~\cite{cao2016deep} combines random surfing with autoencoders.  The model includes three components: random surfing, positive pointwise mutual information (PPMI) calculation, and stacked denoising autoencoders. The random surfing design is used to create a stochastic matrix equivalent to the similarity measure matrix in HOPE~\cite{DBLP:conf/kdd/OuCPZ016} on the input graph. The matrix is transformed into a PPMI matrix and fed into a stacked denoising autoencoder to obtain the embedding. The PPMI matrix input ensures that the autoencoder model captures a higher proximity order. The use of stacked denoising autoencoders also helps make the model robust in the event of noise in the graph, as well as in seizing the internal structure needed for downstream tasks.
	
	\textbf{GAE \& VGAE.}
	Both MLP-based and RNN-based strategies only consider the contextual information and ignore the feature information of the nodes. To encode both, GAE~\cite{kipf2016variational} uses GCN~\cite{DBLP:conf/iclr/KipfW17}. The encoder is in the form,
	\begin{equation}
		\label{vae_enc}
		\operatorname{Enc}({A},{X}) = \operatorname{GraphConv}\left(\sigma (\operatorname{GraphConv} ({A},{X}))\right),
	\end{equation}
	where $\operatorname{GraphConv}(\cdot)$ is a graph convolutional layer defined in~\cite{DBLP:conf/iclr/KipfW17}, $\sigma(\cdot)$ is the activation function and $A$ , $X$ is adjacency matrix and attribute matrix respectively. The decoder of GAE is defined as
	\begin{equation}
		\label{vae_dec}
		\operatorname{Dec}{(\mathbf{z}_{u},\mathbf{z}_{v})} = \mathbf{z}_{u}^T \mathbf{z}_{v}.
	\end{equation}
	Variational GAE (VGAE)~\cite{kipf2016variational} learns about the distribution of the data in which the variation lower bound $\mathcal{L}$ is optimized directly by reconstructing the adjacency matrix.
	\begin{equation}
		\label{vgae}
		\mathcal{L}=\mathbb{E}_{q(\mathbf{Z} \mid {X}, {A})}[\log p({A} \mid \mathbf{Z})]-\operatorname{KL}[q(\mathbf{Z} \mid {X}, {A}) \| p(\mathbf{Z})],
	\end{equation}
	where $\operatorname{KL}[q(\cdot) \| p(\cdot)]$ is the Kullback-Leibler divergence between $q(\cdot)$ and $p(\cdot)$. Moreover, we have
	\begin{equation}
		q(\mathbf{Z} \mid {X}, {A})=\prod_{i=1}^{N} \mathcal{N}\left(\mathbf{z}_{i} \mid {\mu}_{i}, \operatorname{diag}\left({\sigma}_{i}^{2}\right)\right),
	\end{equation}
	and
	\begin{equation}
		p({A} \mid \mathbf{Z})=\prod_{i=1}^{N} A_{ij}\sigma\left(\mathbf{z}_{i}^{\top} \mathbf{z}_{j}\right) + \left(1-A_{ij}\right)\left(1-\sigma\left(\mathbf{z}_{i}^{\top} \mathbf{z}_{j}\right)\right).
	\end{equation}
	
	\textbf{Summary.}
	It should be noted from Eq.~(\ref{autoencoder_embedding_goal}) that the encoder unit does depend on the specific $\mathbf{s}_{i}$ vector, which gives the crucial information relating to the local community structure of $v_{i}$. TABLE~\ref{autoencoder_summary} sums up the main components of these methods, and their architectures are compared as Fig.~\ref{fig:graph_based}.

	\begin{table*}[!ht]
		\centering
		\caption{Summary of AutoEncoder-based Deep Graph Embedding Methods}
		\label{autoencoder_summary}
		\begin{tabular}{cccccc}
			\hline
			\textbf{Method} &
			\textbf{Encoder} &
			\textbf{Decoder} &
			\textbf{Similarity Measure} &
			\textbf{Loss Function} &
			\textbf{Time Complexity} \\
			\hline
			SDNE~\cite{wang2016structural} &
			MLP &
			MLP &
			$\mathbf{s}_{u}$ &
			$\sum_{{u} \in \mathcal{V}}\left\|\operatorname{Dec}\left(\mathbf{z}_{u}\right)-\mathbf{s}_{u}\right\|_{2}^{2}$ &
			${O}(|\mathcal{V} \| \mathcal{E}|)$ \\
			DNGR~\cite{cao2016deep} &
			MLP &
			MLP &
			$\mathbf{s}_{u}$ &
			$\sum_{{u} \in {\mathcal{V}}}\left\|\operatorname{Dec}\left(\mathbf{z}_{u}\right)-\mathbf{s}_{u}\right\|_{2}^{2}$ &
			${O}\left(|\mathcal{V}|^{2}\right)$ \\
			GAE~\cite{kipf2016variational} &
			GCN &
			$\mathbf{z}_{u}^{\top} \mathbf{z}_{v}$ &
			$A_{uv}$ &
			$\sum_{{u} \in {\mathcal{V}}}\left\|\operatorname{Dec}\left(\mathbf{z}_{u}\right)-{A}_{u}\right\|_{2}^{2}$ &
			${O}(|\mathcal{V} \| \mathcal{E}|)$ \\
			VGAE~\cite{kipf2016variational} &
			GCN &
			$\mathbf{z}_{u}^{\top} \mathbf{z}_{v}$ &
			$A_{uv}$ &
			$\mathbb{E}_{q(\mathbf{Z} \mid {X}, {A})}[\log p({A} \mid \mathbf{Z})]-\operatorname{KL}[q(\mathbf{Z} \mid {X}, {A}) \| p(\mathbf{Z})]$ &
			${O}(|\mathcal{V} \| \mathcal{E}|)$ \\
			\hline
		\end{tabular}
	\end{table*}
	
	\subsection{GNN-based methods}
	Several updated, deep embedding strategies are designed to resolve major autoencoder-based method disadvantages by building certain specific functions that rely on the local community of the node but not necessarily the whole graph (Fig.~\ref{fig:graph_based}(5)). The GNN, which is widely used in state-of-the-art deep embedding approaches, can be regarded as a general guideline for the definition of deep neural networks on graphs.

	Like other deep node-level embedding methods, a classifier is trained to predict class labels for the labeled nodes. Then it can be applied to the unlabeled nodes based on the final, hidden state of the GNN-based model. Since GNN consists of two primary operations: the aggregate operation and the update operation, the basic GNN is provided, and then some popular GNN extensions are reviewed with a view to enhancing each process, respectively.
	
	\textbf{Basic GNN.}
	As Gilmer~\etal~\cite{gilmer2017neural} point out, the critical aspect of a basic GNN is that the benefit of \textit{neural message passing} is to exchange and update messages between each node pair by using neural networks.
	
	More specifically, a hidden embedding $\mathbf{h}_{u}^{(k)}$ in each neural message passing through the GNN basic iteration is updated according to message or information from the neighborhood within $\mathcal{N}(u)$ according to each node $u$. This general message can be expressed according to the update rule:
	\begin{equation}\begin{aligned}
			\label{message_passing_update_rule}
			&\mathbf{h}_{u}^{(k+1)} \\
			&=\text { Update }^{(k)}\left(\mathbf{h}_{u}^{(k)}, \text { Aggregate }^{(k)}\left(\left\{\mathbf{h}_{v}^{(k)}, \forall v \in \mathcal{N}(u)\right\}\right)\right) \\
			&=\text { Update }^{(k)}\left(\mathbf{h}_{u}^{(k)}, \mathbf{m}_{\mathcal{N}(u)}^{(k)}\right),
	\end{aligned}\end{equation}
	where
	\begin{equation}
		\mathbf{m}_{\mathcal{N}(u)}^{(k)}=\text { Aggregate }^{(k)}\left(\left\{\mathbf{h}_{v}^{(k)}, \forall v \in \mathcal{N}(u)\right\}\right).
	\end{equation}
	
	It is worth noting that the functions $\textbf{Update}$ and $\textbf{Aggregate}$ must generally be differentiable in Eq.~(\ref{message_passing_update_rule}). The new state is generated in accordance with Eq.~(\ref{message_passing_update_rule}) when the neighborhood message is combined with the previous hidden embedding state. After a number of iterations, the last hidden embedding state converges so that each node's final status is created as the output. Formally, we have, $\mathbf{z}_{u}=\mathbf{h}_{u}^{(K)}, \forall u \in \mathcal{V}$.
	
	The basic GNN model is introduced before reviewing many other GNN-based methods designed to perform SSL tasks. The basic version of GNN aims to simplify the original GNN model, proposed by Scarselli~\etal~\cite{scarselli2008graph}.
	
	The basic GNN message passing is defined as:
	\begin{equation}
		\label{basic_gnn}\mathbf{h}_{u}^{(k)}=\sigma\left(\mathbf{W}_{\text {self }}^{(k)} \mathbf{h}_{u}^{(k-1)}+\mathbf{W}_{\text {neigh }}^{(k)} \sum_{v \in \mathcal{N}(u)} \mathbf{h}_{v}^{(k-1)}+\mathbf{b}^{(k)}\right),\end{equation}
	where $\mathbf{W}_{\text {self }}^{(k)}, \mathbf{W}_{\text {neigh }}^{(k)}$ are trainable parameters and $\sigma$ is the activation function.
	In principle, the messages from the neighbors are summarized first. Then, the neighborhood information and the previously hidden node results are integrated by an essential linear combination. Finally, the joint information uses a nonlinear activation function. It is worth noting that the GNN layer can be easily stacked together following Eq.~(\ref{basic_gnn}). The last layer's output in the GNN model is regarded as the final node embedding result to train a classifier for the downstream SSL tasks.
	
	As previously mentioned, GNN models have all sorts of variants that try to some degree boost their efficiency and robustness. All of them, though, obey the Eq.~(\ref{message_passing_update_rule}) neural message passing structure, regardless of the GNN version explored previously.
	
	\textbf{GCN.}
	As mentioned above, the most simple neighborhood aggregation operation only calculates the sum of the neighborhood encoding states. The critical issue with this approach is that nodes with a large degree appear to derive a better benefit from more neighbors compared to those with a lower number of neighbors.
	
	One typical and straightforward  approach to this problem is to normalize the aggregation process depending on the central node degree. The most popular method  is to use the following symmetric normalization Eq.~(\ref{gcn_neighbor_norm}) employed by Kipf~\etal~\cite{DBLP:conf/iclr/KipfW17} in the graph convolutional network (GCN) model as Eq.~(\ref{gcn_neighbor_norm}).
	\begin{equation}
		\label{gcn_neighbor_norm}
		\mathbf{m}_{\mathcal{N}(u)}=\sum_{v \in \mathcal{N}(u)} \frac{\mathbf{h}_{v}}{\sqrt{|\mathcal{N}(u)| \mid \mathcal{N}(v)} \mid}
	\end{equation}
	
	GCN fully uses the uniform neighborhood grouping technique. Consequently, the GCN model describes the update function as Eq.~(\ref{GCN}). As it is indirectly specified in the update function, no aggregation operation is defined.
	
	\begin{equation}
		\label{GCN}
		\mathbf{h}_{u}^{(k)}=\sigma\left(\mathbf{W}^{(k)} \sum_{v \in \mathcal{N}(u) \cup\{u\}} \frac{\mathbf{h}_{v}}{\sqrt{|\mathcal{N}(u)||\mathcal{N}(v)|}}\right)
	\end{equation}
	
	A vast range of GCN variants is available to boost SSL performance from various aspects \cite{DBLP:conf/nips/Wang0C0PW20,DBLP:conf/nips/FengZDHLXYK020}. Li~\etal~\cite{li2018deeper} were the first to have a detailed insight into the performance and lack of GCN in SSL tasks. Subsequently, GCN extensions for SSL started to propagate~\cite{DBLP:conf/iclr/LiaoBTGUZ18}~\cite{zhang2019bayesian}~\cite{vashishth2019confidence}~\cite{10.1145/3178876.3186116}~\cite{hu2019hierarchical}.
	
	\textbf{GAT.}
	In addition to more general ways of set aggregation, another common approach for improving the aggregation layer of GNNs is to introduce certain attention mechanisms~\cite{BahdanauCB14}. The basic theory is to give each neighbor a weight or value of significance which is used to weight the influence of this neighbor during the aggregation process. The first GNN to use this focus was Cucurull~\etals Graph Attention Network (GAT), which uses attention weights to describe a weighted neighboring amount:
	\begin{equation}\mathbf{m}_{\mathcal{N}(u)}=\sum_{v \in \mathcal{N}(u)} \alpha_{u, v} \mathbf{h}_{v},\end{equation}
	where $\alpha_{u, v}$ denotes the attention on neighbor $v \in \mathcal{N}(u)$ when we are aggregating information at node $u$. In the original GAT paper, the attention weights are defined as
	\begin{equation}\alpha_{u, v}=\frac{\exp \left(\mathbf{a}^{\top}\left[\mathbf{W} \mathbf{h}_{u} \oplus \mathbf{W} \mathbf{h}_{v}\right]\right)}{\sum_{v^{\prime} \in \mathcal{N}(u)} \exp \left(\mathbf{a}^{\top}\left[\mathbf{W h}_{u} \oplus \mathbf{W h}_{v^{\prime}}\right]\right)},\end{equation}
	where $\mathbf{a}$ is a trainable attention vector, $\mathbf{W}$ is a trainable matrix, and denotes the concatenation operation.
	
	\textbf{GraphSAGE.}
	%\subsection{Generalized update operation}
	\label{Generalized update operation}
	Over-smoothing is an obvious concern for GNN. Over-smoothing after several message iterations is almost unavoidable as the node-specific information is ``washed away." The use of vector concatenations or skip connections, which both preserve information directly from the previous rounds of the update, is one fair way to lessen this concern. For general purposes, $\text{Update}_{\text{base}}$ denotes a basic update rule.
	
	One of the simplest updates in GraphSage~\cite{hamilton2017inductive} for skip connection uses a concatenation vector to contain more node-level information during a messages passage process:
	\begin{equation}\text {Update}\left(\mathbf{h}_{u}, \mathbf{m}_{\mathcal{N}}(u)\right)=\left[\text {Update}_{\text {base}}\left(\mathbf{h}_{u}, \mathbf{m}_{\mathcal{N}(u)}\right) \oplus \mathbf{h}_{u}\right],\end{equation}
	where the output from the basic update function is concatenated with the previous layer representation of the node. The critical observation is that this designed model is encouraged to detach information during the message passing operation.
	
	\textbf{GGNN.}
	Parallel to the above work, researchers also are motivated by the approaches taken by recurrent neural networks (RNNs) to improve stability. One way to view the GNN message propagation algorithm is to gather an observation from neighbors of the aggregation process and then change each node's hidden state. In this respect, specific approaches can be used explicitly based on observations to check the hidden states of RNN architectures.
	
	For example, one of the earliest GNN variants which put this idea into practice is proposed by Li~\etal~\cite{gatedGNN}, in which the update operation is defined as Eq.~(\ref{gatedGNN})
	\begin{equation}
		\label{gatedGNN}
		\mathbf{h}_{u}^{(k)}=\operatorname{GRU}\left(\mathbf{h}_{u}^{(k-1)}, \mathbf{m}_{\mathcal{N}(u)}^{(k)}\right),\end{equation}
	where GRU is a gating mechanism function in recurrent neural networks, introduced by Kyunghyun Cho~\etal~\cite{chung2014empirical}. Another related approach would be similar improvements based on the LSTM architecture~\cite{lstmGNN}.
	
	\textbf{Summary.} The main point of graph-based models for DSSL is to perform label inference on a constructed similarity graph so that the label information can be propagated from the labeled samples to the unlabeled ones by incorporating both the topological and feature knowledge. Moreover, the involvement of deep learning models in GSSL helps generate more discriminative embedding representations that are beneficial for the downstream SSL tasks, thanks to the more complex encoder functions.

	\section{Pseudo-labeling methods}\label{sec:pseudoLabeling}
	The pseudo-labeling methods differ from the consistency regularization methods in that the consistency regularization methods usually rely on consistency constraint of rich data transformations.  In contrast, pseudo-labeling methods rely on the high confidence of pseudo-labels, which can be added to the training data set as labeled data. 
	There are two main patterns, one is to improve the performance of the whole framework based on the disagreement of views or multiple networks, and the other is self-training, in particular, the success of self-supervised learning in unsupervised domain makes some self-training self-supervised methods realized.
	
	%fig: pseudo-labeling models
	\begin{figure*}[!t]
		\centering
		\includegraphics[width=7.0in]{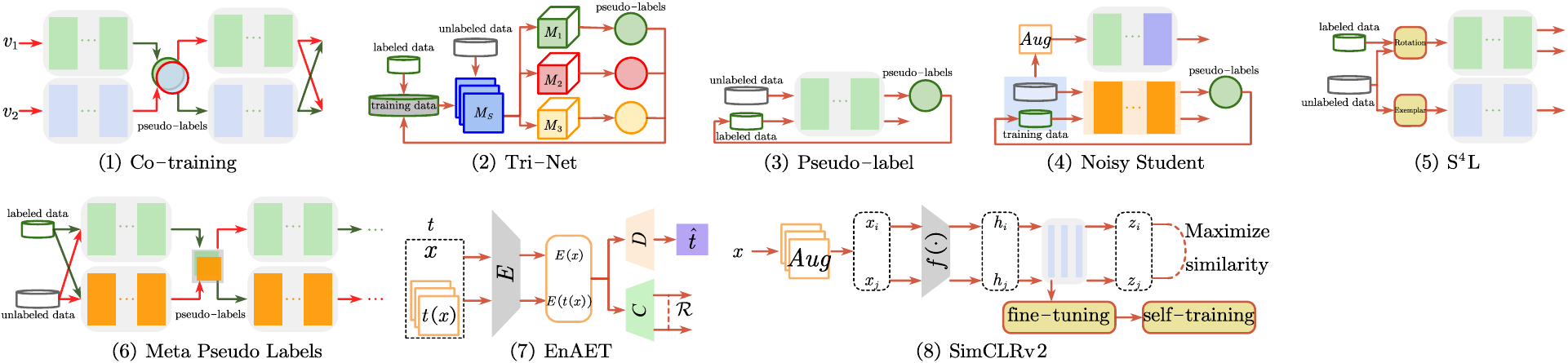}
		\caption{A glimpse of the diverse range of architectures used for pseudo-label semi-supervised methods. The same color and structure have the same meaning as shown in Figure \ref{fig:consistencyRegularization}. $M_s$ denotes shared module, $M_1,M_2$ and $M_3$ are three different modules in Tri-Net. ``Rotation " and ``Exemplar"
			represent $S^4L$-Rotation and $S^4L$-Exemplar, respectively.   }
		\label{fig:pseudoLabel}
	\end{figure*}
	
	\subsection{Disagreement-based models}\label{sec:disagreement}
	The idea of disagreement-based SSL is to train multiple learners for the task and exploit the disagreement during the learning process  \cite{DBLP:journals/kais/ZhouL10}. In such model designs, two or three different networks are trained simultaneously and label unlabeled samples for each other.
	% Disagreement-based methods differ in whether the data has multiple views \ie, Co-training \cite{DBLP:conf/colt/BlumM98,DBLP:conf/eccv/QiaoSZWY18}  for multiview data or Tri-Net \cite{DBLP:conf/ijcai/ChenWGZ18} for single-view data.
	
	\textbf{Deep co-training.}
	Co-training \cite{DBLP:conf/colt/BlumM98} framework assumes each data $x$ in the dataset has two different and complementary views, and each view is sufficient for training a good classifier.
	Because of this assumption, Co-training learns two different classifiers on these two views (see Fig.~\ref{fig:pseudoLabel}(1)).
	Then the two classifiers are applied to predict each view's unlabeled data and label the most confident candidates for the other model.
	This procedure is  iteratively repeated till unlabeled data are exhausted, or some condition is met (such as the maximum number of iterations is reached). Let $v_1$ and $v_2$ as two different views of data such that $x =(v_1,v_2)$. Co-training assumes that $\mathcal{C}_1$ as the classifier trained on View-$1$ $v_1$ and $\mathcal{C}_2$ as the classifier trained on View-$2$ $v_2$ have consistent predictions on $\mathcal{X}$. 
	In the objective function, the Co-training assumption can be model as:
	\begin{equation}
		\mathcal{L}_{ct}=H(\frac{1}{2}(\mathcal{C}_1(v_1)+\mathcal{C}_2(v_2)))-\frac{1}{2}(H(\mathcal{C}_1(v_1))+H(\mathcal{C}_2(v_2))),
	\end{equation}
	where $H(\cdot)$ denotes the entropy, the Co-training assumption is formulated as $\mathcal{C}(x)=\mathcal{C}_1(v_1)=\mathcal{C}_2(v_2), \forall x=(v_1,v_2)\sim \mathcal{X}$. On the labeled dataset $X_L$, the supervised loss function can be the standard cross-entropy loss
	\begin{equation}
		\mathcal{L}_{s}=H(y,\mathcal{C}_1(v_1))+H(y,\mathcal{C}_2(v_2)),
	\end{equation}
	where $H(p,q)$ is the cross-entropy between distribution $p$ and $q$.
	
	The key to the success  of Co-training is that the two views are different and complementary. However,  the loss function $\mathcal{L}_{ct} $ and $\mathcal{L}_s$ only ensure the model tends to be consistent for the predictions on the dataset. To address this problem, \cite{DBLP:conf/eccv/QiaoSZWY18} forces to add the View Difference Constraint to the previous Co-training model, and formulated as:
	\begin{equation}
		\exists \mathcal{X}': \mathcal{C}_1(v_1) \neq \mathcal{C}_2(v_2), \forall x=(v_1,v_2)\sim \mathcal{X}',
	\end{equation}
	where $\mathcal{X}'$ denotes the adversarial examples of $\mathcal{X}$, thus $\mathcal{X}' \cap \mathcal{X}=\emptyset$.
	In the loss function, the View Difference Constraint can be model by minimizing the cross-entropy between $\mathcal{C}_2(x)$ and $\mathcal{C}_1(g_2(x))$, where $g(\cdot)$ denotes the adversarial examples generated by the generative model. Then, this part loss function is:
	\begin{equation}
		\mathcal{L}_{dif}(x)=H(\mathcal{C}_1(x),\mathcal{C}_2(g_1(x)))+H(\mathcal{C}_2(x),p_1(g_2(x))).
	\end{equation}
	%In Deep Co-training \cite{DBLP:conf/eccv/QiaoSZWY18}, the objective function can be formed as:
	%\begin{equation}
	%	\mathcal{L}=\mathbb{E}_{(x,y)\in X_L }\mathcal{L}_{s}(x,y)
	%	+\alpha \mathbb{E}_{x\in X_U}\mathcal{L}_{ct}(x)
	%	+\beta\mathbb{E}_{x\in X}\mathcal{L}_{dif}(x),
	%\end{equation}
	%where $\alpha$ and $\beta$ are linear hyper-parameters.
	
	Some other research works also explore to apply co-training into neural network model training. For example, \cite{DBLP:conf/ijcai/ChengZCLHR16} treats the RGB and depth of an image as two independent views for object recognition. Then, co-training is performed to train two networks on the two views. Next, a fusion layer is added to combine the two-stream networks for recognition, and the overall model is jointly trained. Besides, in sentiment classification, \cite{DBLP:conf/acl/XiaWDL15} considers the original review and the automatically constructed anonymous review as two opposite sides of one review and then apply the co-training algorithm. One crucial property of  \cite{DBLP:conf/acl/XiaWDL15} is that two views are opposing and therefore associated with opposite class labels.
	
	\textbf{Tri-Net.}
	Tri-net \cite{DBLP:conf/ijcai/ChenWGZ18}, a deep learning-based method inspired by the tri-training \cite{DBLP:journals/tkde/ZhouL05}. The tri-training learns three classifiers from three different training sets, which are obtained by utilizing bootstrap sampling. The framework (as shown in Fig.~\ref{fig:pseudoLabel}(2)) of tri-net can be intuitively described as follows.
	Output smearing \cite{DBLP:journals/ml/Breiman00} is used to add random noise to the labeled sample to generate different training sets and help learn three initial modules. The three models then predict the pseudo-label for unlabeled data.  With the predictions of two modules for unlabeled instances consistent, the pseudo-label is considered to be confident and stable. The labeled sample is added to the training set of the third module, and then the third module is fine-tuned on the augmented training set. During the augmentation process, the three modules become more and more similar, so the three modules are fine-tuned on the training sets respectively to ensure diversity. Formally, the output smearing is used to construct three different training sets $\{\mathcal{L}_{os}^j=(x_i, \hat{y}_i^j), j=1, 2, 3\}$  from the initial labeled set $X_L$. Then tri-net can be initialized by minimizing the sum of standard softmax cross-entropy loss function from the three training sets, 
	\begin{align}
		\mathcal{L}=&\frac{1}{L}\sum_{i=1}^L \left\{ \mathcal{L}_y(M_1(M_S(x_i)),\hat{y}_i^1)
		+ \mathcal{L}_y(M_2(M_S(x_i)),\hat{y}_i^2)\nonumber\right.\\
		& \left.+ \mathcal{L}_y(M_3(M_S(x_i)),\hat{y}_i^3)\right\},
	\end{align}
	where $\mathcal{L}_y$ is the standard softmax cross-entropy loss function; $M_S$ denote a shared module, and $M_1,M_2,M_3$ is the three different modules; $M_j(M_S(x_i)), j=1,2,3$ denotes the outputs of the shared features generated by $M_S$.
	In the whole procedure, the unlabeled sample can be pseudo-labeled  by the maximum posterior probability,
	\begin{equation}
		\begin{aligned}
			y=&\argmax_{k\in \{1,2,\dots, K\}}\left\{p(M_1(M_S(x))=k|x)+\nonumber\right.\\
			&\left. p(M_2(M_S(x))=k|x)+p(M_3(M_S(x))=k|x) \right\}.
		\end{aligned}
	\end{equation}

	\textbf{Summary.} The disagreement-based SSL methods exploit the unlabeled data by training multiple learners, and the ``disagreement"
	among these learners is crucial. When the data has two sufficient redundancy and conditional independence views, Deep Co-training  \cite{DBLP:conf/eccv/QiaoSZWY18} improves the disagreement by designing a View Difference Constraint. Tri-Net \cite{DBLP:conf/ijcai/ChenWGZ18} obtains three labeled datasets by bootstrap sampling and trains three different learners. These methods in this category are less affected by model assumptions, non-convexity of the loss function and the scalability of the learning algorithms.

	\subsection{Self-training models }\label{sec:pLabel}
	Self-training algorithm leverages the model's own confident predictions to produce the pseudo labels for unlabeled data. In other words, it can add more training data by using existing labeled data to predict the labels of unlabeled data.
	%Because of its simplicity and generality, self-training is successfully applied in various tasks such as Named Entity Recognition \cite{DBLP:conf/aaai/HeS17}, contour detection \cite{DBLP:conf/cvpr/ZhangXSY16}, machine translation \cite{DBLP:conf/emnlp/OrorbiaGR15}, and object detection \cite{DBLP:conf/cvpr/MisraSH15}. Specifically, \cite{DBLP:conf/aaai/HeS17} chooses the most confident named entity recognition predictions of the unlabeled data as the additional targets to boost the performance.  \cite{DBLP:conf/emnlp/OrorbiaGR15} first trains the neural translation model on the parallel corpus and then uses the learned model to translate a monolingual corpus, wherein the monolingual corpus and its translations constitute an additional pseudo-parallel corpus.
	
	\textbf{EntMin.}
	Entropy Minimization (EntMin) \cite{DBLP:conf/nips/GrandvaletB04} is a method of entropy regularization, which can be used to realize SSL by encouraging the model to make low-entropy predictions for unlabeled data and then using the unlabeled data in a standard supervised learning setting. In theory, entropy minimization can prevent the decision boundary from passing through a high-density data points region, otherwise it would be forced to produce low-confidence predictions for unlabeled data.
	
	\textbf{Pseudo-label.}
	Pseudo-label \cite{Lee2013PseudoLabelT} proposes a simple and efficient formulation of training neural networks in a semi-supervised fashion, in which the network is trained in a supervised way with labeled and unlabeled data simultaneously. As illustrated in Fig.~\ref{fig:pseudoLabel}(3), the model is trained on labeled data in a usual supervised manner with a cross-entropy loss. For unlabeled data, the same model is used to get predictions for a batch of unlabeled samples. The maximum confidence prediction is called a pseudo-label, which has the maximum predicted probability.
	
	That is, the pseudo-label model trains a neural network with the loss function $\mathcal{L}$, where:
	\begin{equation}
		\mathcal{L}=\frac{1}{n}\sum_{m=1}^{n}\sum_{i=1}^{K}\mathcal{R}(y_i^m, f_i^m)+\alpha (t) \frac{1}{n'}\sum_{m=1}^{n'}\sum_{i=1}^{K}\mathcal{R}(y_i^{'m},f_i^{'m}),
	\end{equation}
	where $n$ is the number of mini-batch in labeled data for SGD, $n'$ for unlabeled data, $f_i^m$ is the output units of $m$'s sample in labeled data, $y_i^m$ is the label of that, $y_i^{'m}$ for unlabeled data, $y_i^{'m}$ is the pseudo-label of that for unlabeled data, $\alpha(t)$ is a coefficient balancing the supervised and unsupervised loss terms.
	
	\textbf{Noisy Student.}
	Noisy Student \cite{DBLP:conf/cvpr/XieLHL20} proposes a semi-supervised method inspired by knowledge distillation \cite{DBLP:journals/corr/HintonVD15} with equal-or-larger student models.
	The framework is shown in Fig.~\ref{fig:pseudoLabel}(4). The teacher EfficientNet \cite{DBLP:conf/icml/TanL19} model is first trained
	on labeled images to generate pseudo labels for unlabeled examples. Then a larger EfficientNet model as a student is trained on the combination of labeled and pseudo-labeled examples. These combined instances are augmented using data augmentation techniques such as RandAugment \cite{DBLP:conf/cvpr/CubukZSL20}, and model noise such as Dropout and stochastic depth are also incorporated in the student model during training. After a few iterations of this algorithm, the student model becomes the new teacher to relabel the unlabeled data and this process is repeated.
	
	\textbf{$\bm{S^4L}$.}
	Self-supervised Semi-supervised Learning ($S^4L$) \cite{DBLP:conf/iccv/BeyerZOK19} tackles the problem of SSL by employing self-supervised learning \cite{DBLP:conf/cvpr/KolesnikovZB19} techniques to learn useful representations from the image databases. The architecture of $S^4L$ is shown in Fig.~\ref{fig:pseudoLabel}(5). The conspicuous self-supervised techniques are predicting image rotation \cite{DBLP:conf/iclr/GidarisSK18} and exemplar \cite{DBLP:conf/nips/DosovitskiySRB14,DBLP:journals/pami/DosovitskiyFSRB16}. Predicting image rotation is a pretext task that anticipates an angle of the rotation transformation applied to an input example. In $S^4L$, there are four rotation degrees $\{0^{\circ}, 90^{\circ} , 180^{\circ}, 270^{\circ}\}$ to rotate input images. The $S^4L$-Rotation loss is the cross-entropy loss on outputs predicted by those rotated images. $S^4L$-Exemplar introduces an exemplar loss which encourages the model to learn a representation that is invariant to heavy image augmentations. Specifically, eight different instances of each image are produced by inception cropping \cite{DBLP:conf/cvpr/SzegedyLJSRAEVR15}, random horizontal mirroring, and HSV-space color randomization as in  \cite{DBLP:conf/nips/DosovitskiySRB14}. Following \cite{DBLP:conf/cvpr/KolesnikovZB19}, the loss term on unsupervised images uses the batch hard triplet loss \cite{DBLP:journals/corr/HermansBL17} with a soft margin.
	
	\textbf{MPL.}
	In the SSL, the target distributions are often generated on unlabeled data by a shaped teacher model trained on labeled data. The constructions for target distributions are heuristics that are designed prior to training, and cannot adapt to the learning state of the network training. Meta Pseudo Labels (MPL) \cite{DBLP:journals/corr/abs-2003-10580} designs  a teacher model that assigns distributions to input examples to train the student model. Throughout the course of the student's training, the teacher observes the student's performance on a held-out validation set, and learns to generate target distributions so that if the student learns from such distributions, the student will achieve good validation performance. The training procedure of MPL  involves two alternating processes. As shown in Fig.~\ref{fig:pseudoLabel}(6), the teacher $g_{\phi}(\cdot)$ produces the conditional class distribution $g_{\phi}(x)$ to train the student. The pair $(x,g_{\phi}(x))$ is then fed into the student network $f_{\theta}(\cdot)$ to update its parameters $\theta$ from the cross-entropy loss. After the student network updates its parameters, the model evaluates  the new parameters $\theta'$ based on the samples from the held-out validation dataset. Since the new parameters of the student depend on the teacher, the dependency allows us to compute the gradient of the loss to update the teacher's parameters.\
	
	%self-supervised Wasserstein pseudo-labeling for semi-supervised image classification.

	\textbf{EnAET.}  Different from the previous semi-supervised methods and $S^4L$ \cite{DBLP:conf/iccv/BeyerZOK19}, EnAET 
	\cite{DBLP:journals/tip/WangKLQ21} trains an Ensemble of Auto-Encoding Transformations to enhance the learning ability of the model. 
	%The EnAET pipeline is shown in Fig.~\ref{fig:pseudoLabel}(7). 
	The core part of this framework is that EnAET integrates an ensemble of spatial and non-spatial transformations to self-train a good feature representation \cite{DBLP:conf/cvpr/ZhangQWL19}. EnAET incorporates four spatial transformations and a combined non-spatial transformation. The spatial transformations are Projective transformation, Affine transformation, Similarity transformation and Euclidean transformation.  The non-spatial transformation is composed of different colors, contrast,  brightness and  sharpness with four strength parameters. As shown in Fig.~\ref{fig:pseudoLabel}(7), EnAET learns an encoder $E: x\rightarrow E(x), t(x)\rightarrow E(t(x))$ on an original instance and its transformations. Meanwhile, a decoder $D:[E(x), E(t(x))]\rightarrow \hat{t}$ is learned to estimate $\hat{t}$ of input transformation. Then, we can get an AutoEncoding Transformation (AET) loss,
	\begin{equation}
		\mathcal{L}_{AET}=\mathbb{E}_{x,t(x)}\|D[E(x),E(t(x))]-t(x)\|^2,
	\end{equation}
	and EnAET adds the AET loss to the SSL loss as a regularization term. Apart from the AET loss, EnAET explore a pseudo-labeling consistency by minimizing the KL divergence between $P(y|x)$ on an original sample $x$ and $P(y|t(x))$ on a transformation $t(x)$.
	
	\textbf{SimCLRv2.} 
	SimCLRv2 \cite{DBLP:conf/nips/ChenKSNH20} modifies 
	SimCLR \cite{DBLP:conf/icml/ChenK0H20} for SSL problems. Following the paradigm of supervised fine-tuning after unsupervised pretraining, SimCLRv2 uses unlabeled samples in a task-agnostic way, and shows that a big (deep and wide) model can surperisingly effective for semi-supervised learning. As shown in Fig.~\ref{fig:pseudoLabel}(8), the SimCLRv2 can be summarized in three steps: unsupervised or self-supervised pre-training, supervised fine-tuning on 1\% or 10\% labeled samples, and self-training with task-specific unlabeled examples. In the pre-training step, SimCLRv2 learns representations by maximizing the contrastive learning loss function in latent space, and the loss function is constructed based on the consistency of different augmented views of the same example. The contrastive loss is
	\begin{equation}
		\ell_{i,j}=-\log \frac{\exp(sim(z_i, z_j)/\tau)}{\sum_{k=1}^{2N}\mathbb{I}_{[k\neq i]}\exp (sim(z_i,z_k)/\tau)},
	\end{equation}
	where $(i,j)$ is a pair of positive example augmented from the same sample. $sim(\cdot, \cdot)$ is cosine similarity, and $\tau$ is a temperature parameter. In the self-training step, SimCLRv2 uses task-specific unlabeled samples, and the fine-tuning network acts as a teacher model to minimize the following distillation loss,
	\begin{equation}
		\mathcal{L}^{\text{distill}}=-\sum_{x_i \in X}\big[\sum_y P^T(y|x_i; \tau)\log P^S(y|x_i;\tau)\big],
	\end{equation}
	where $P^T(y|x_i;\tau)$ and $P^S(y|x_i;\tau)$ are produced by teacher network and student network, respectively.

	\textbf{Summary.} In general, self-training is a way to get more training data by a series of operations to get pseudo-labels of unlabeled data. Both EntMin \cite{DBLP:conf/nips/GrandvaletB04} and Pseudo-label \cite{Lee2013PseudoLabelT} use entropy minimization to get the pseudo-label with the highest confidence as the ground truth for unlabeled data. Noisy Student \cite{DBLP:conf/cvpr/XieLHL20} utilizes a variety of techniques when training the student network, such as data augmentation, dropout and stochastic depth. $S^4L$ \cite{DBLP:conf/iccv/BeyerZOK19} not only uses data augmentation techniques, but also adds another $4$-category task to improve model performance. MPL \cite{DBLP:journals/corr/abs-2003-10580} modifies Pseudo-label  \cite{Lee2013PseudoLabelT} by deriving the teacher network's update rule  from the feedback of the student network. Emerging techniques (\eg, rich data augmentation strategies, meta-learning, self-supervised learning ) and network architectures (\eg, EfficientNet \cite{DBLP:conf/icml/TanL19}, SimCLR \cite{DBLP:conf/icml/ChenK0H20}) provide powerful support for the development of self-training methods.

	\section{Hybrid methods}\label{sec:hybridModel}
	Hybrid methods combine ideas from the above-mentioned methods  such as pseudo-label, consistency regularization and entropy minimization for performance improvement. Moreover, a learning principle, namely Mixup \cite{DBLP:conf/iclr/ZhangCDL18}, is introduced in those hybrid methods. It can be considered as a simple, data-agnostic data augmentation approach, a convex combination of paired samples and their respective labels. Formally, Mixup constructs virtual training examples,
	\begin{equation}
		\tilde{x}=\lambda x_i+(1-\lambda)x_j,\quad
		\tilde{y}=\lambda y_i+(1-\lambda)y_j,
	\end{equation}
	where $(x_i,y_i)$ and $(x_j,y_j)$ are two instances from the training data, and $\lambda \in [0,1]$. 
	Therefore, Mixup extends the training data set by a hard constraint  that linear interpolations of samples should lead to the linear interpolations of the corresponding labels.
	
	\begin{figure*}[!t]
		\centering
		\includegraphics[width=7.0in]{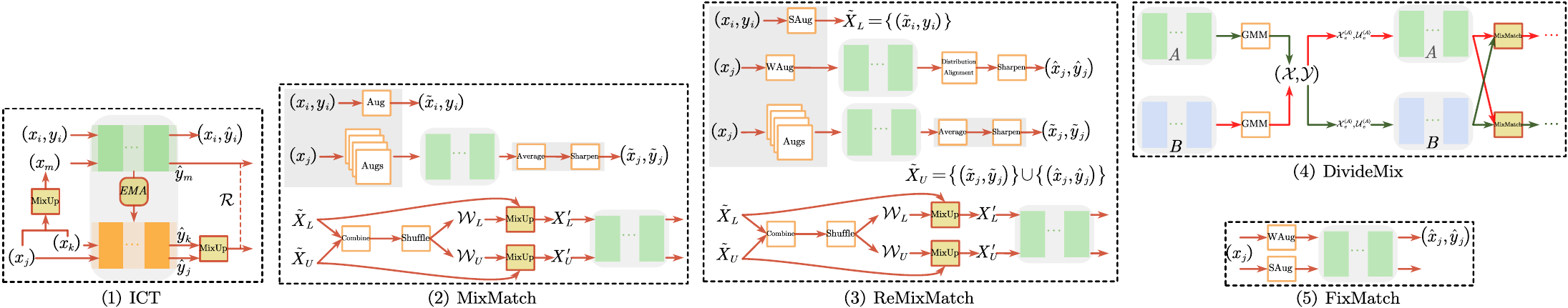}
		\caption{A glimpse of the diverse range of architectures used for hybrid semi-supervised methods. ``Mixup"
			is Mixup operator \cite{DBLP:conf/iclr/ZhangCDL18}. ``MixMatch"
			is MixMatch \cite{DBLP:conf/nips/BerthelotCGPOR19} in figure (2). ``GMM"
			is short for Gaussian Mixture Model. ``SAug" and ``WAug"
			represent 
			Strong Augmentation and Weak Augmentation, respectively. }
		\label{fig:hybridModel}
	\end{figure*}

	\textbf{ICT.}
	Interpolation Consistency Training (ICT) \cite{DBLP:conf/ijcai/VermaLKBL19} regularizes SSL by encouraging the prediction at an interpolation of two unlabeled examples to be consistent with the interpolation of the predictions at those points. The architecture is  shown in Fig.\ref{fig:hybridModel}(1). The low-density separation assumption inspires this method, and Mixup can achieve broad margin decision boundaries. It is done by training the model $f(\theta)$ to predict $\text{Mix}_{\lambda}(\hat{y}_j,\hat{y}_k)$ at location $\text{Mix}_{\lambda}(x_j,x_k)$ with the Mixup operation: $\text{Mix}_{\lambda}(a,b)=\lambda a+(1-\lambda) b$. In semi-supervised settings, ICT extends Mixup by training the model $f(\theta, x)$ to predict the ``fake label" $\text{Mix}_{\lambda}(f(\theta ,x_j),f(\theta ,x_k))$ at location $\text{Mix}_{\lambda}(x_j,x_k)$. Moreover, the model $f_{\theta}$ predict the fake label $\text{Mix}_{\lambda}(f_{\theta '}(x_i),f_{\theta '}(x_j))$ at location $\text{Mix}_{\lambda}(x_i,x_j)$, where $\theta'$ is a moving average of $\theta$, like  \cite{DBLP:conf/nips/TarvainenV17}, for a more conservation consistent regularization. Thus, the ICT term is
	\begin{equation}
		\mathbb{E}_{x\in X}\mathcal{R}(f(\theta, \text{Mix}_{\lambda}(x_i,x_j)), \text{Mix}_{\lambda}(f(\theta', x_i),f(\theta',x_j)).
	\end{equation}
	
	\textbf{MixMatch.}
	MixMatch \cite{DBLP:conf/nips/BerthelotCGPOR19} combines consistency regularization and entropy minimization in a unified loss function. This model operates by producing pseudo-labels for each unlabeled instance and then training the original labeled data with the pseudo-labels for the unlabeled data using fully-supervised techniques. The main aim of this algorithm is to create the collections  $X'_L$ and $X'_U$, which are made up of augmented labeled and unlabeled samples that were generated using Mixup. Formally, MixMatch produces an augmentation of each labeled instance $(x_i,y_i)$ and $K$ weakly augmented version of each unlabeled instance $(x_j)$ with $k \in \{1,\ldots, K\}$. Then, it generates a pseudo-label $\bar{y}_j$ for each $x_j$ by computing the average prediction across the $K$ augmentations. The pseudo-label distribution is then sharpened by adjusting temperature scaling to get the final pseudo-label $\tilde{y}_j$. After the data augmentation, the batches of augmented labeled  examples and unlabeled with pseudo-label  examples are combined,  then the whole group is shuffled. This group is divided into two parts:  the first $L$ samples were taken as $\mathcal{W}_L$, and the remaining taken as $\mathcal{W}_U$. The group $\mathcal W_L$ and the augmented labeled batch $\tilde{X}_L$ are fed into the Mixup algorithm to compute examples $(x',y')$ where $x'=\lambda x_1+(1-\lambda)x_2$ for $\lambda \sim \text{Beta}(\alpha, \alpha)$. Similarly, Mixup is applied between the remaining $\mathcal{W}_U$ and the augmented unlabeled group $\tilde{X}_U$. MixMatch conducts traditional fully-supervised training with a standard cross-entropy loss for a supervised dataset and a mean square error for unlabeled data given these mixed-up samples.
	
	\textbf{ReMixMatch.}
	ReMixMatch \cite{DBLP:conf/iclr/BerthelotCCKSZR20} extends MixMatch \cite{DBLP:conf/nips/BerthelotCGPOR19} by introducing distribution alignment and augmentation anchoring. Distribution alignment encourages the marginal distribution of aggregated class predictions on unlabeled data close to the marginal distribution of ground-truth labels. Augmentation anchoring replaces the consistency regularization component of MixMatch. This technique generates  multiple strongly augmented versions of input and encourages each output to be close to predicting a weakly-augmented variant of the same input. A variant of AutoAugment \cite{DBLP:conf/cvpr/CubukZMVL19} dubbed ``CTAugment" is also proposed to produce strong augmentations, which  learns the augmentation policy alongside the model training.  As the procedure of ReMixmatch presented in Fig.~\ref{fig:hybridModel}(3), an ``anchor"
	is generated by applying weak augmentation to a given unlabeled example and then $K$ strongly-augmented versions of the same unlabeled example using CTAugment.
	
	\textbf{DivideMix.}
	DivideMix \cite{DBLP:conf/iclr/LiSH20} presents a new SSL framework to handle the problem of learning with noisy labels. As shown in Fig.~\ref{fig:hybridModel}(4), DivideMix proposes co-divide, a process that trains two networks simultaneously. For each network, a dynamic Gauss Mixed Model (GMM) is fitted on the loss distribution of each sample to divide the training set into labeled data and unlabeled data. The separated data sets are then used to train the next epoch's networks. In the follow-up SSL process, co-refinement and co-guessing are used to improve MixMatch \cite{DBLP:conf/nips/BerthelotCGPOR19} and solve the problem of learning with noisy labels.
	
	\textbf{FixMatch.}
	FixMatch \cite{DBLP:conf/nips/SohnBCZZRCKL20} combines consistency regularization and pseudo-labeling while vastly simplifying the overall method.
	The key innovation comes from the combination of these two ingredients, and the use of a separate weak and strong augmentation in the consistency regularization approach. Given an instance, only when the model predicts a high-confidence label can the predicted pseudo-label be identified as ground-truth. As shown in Fig.\ref{fig:hybridModel}(5), given an instance $x_j$, FixMatch firstly  generates pseudo-label $\hat{y}_j$ for weakly-augmented unlabeled instances $\hat{x}_j$.  Then, the model trained on the weakly-augmented samples is used to predict pseudo-label in the strongly-augmented version of $x_j$. 
	In FixMatch, weak augmentation is a standard flip-and-shift augmentation strategy, randomly flipping images horizontally with a probability. For strong augmentation, there are two approaches which are based on \cite{DBLP:conf/cvpr/CubukZMVL19}, \ie, RandAugment \cite{DBLP:conf/cvpr/CubukZSL20} and CTAugment \cite{DBLP:conf/iclr/BerthelotCCKSZR20}. Moreover, Cutout \cite{DBLP:journals/corr/abs-1708-04552} is followed by either of these strategies.

	%AlphaMatch: Improving Consistency for semi-supervised learning with alpha-divergence.

	\textbf{Summary.}
	As discussed above, the hybrid methods unit the most successful approaches in SSL, such as pseudo-labeling, entropy minimization and consistency regularization,  and adapt them to achieve state-of-the-art performance. In TABLE~\ref{tab:augmentation}, we summarize some techniques that can be used in consistency training to improve model performance.

	\begin{table}
		\caption{Summary of Input Augmentations and Neural Network Transformations}
		\label{tab:augmentation}
		\centering
		\begin{tabular}{>{\centering\arraybackslash}m{1.8cm}|>{\centering\arraybackslash}m{1.7cm}|>{\centering\arraybackslash}m{4cm}}
			\hline
			& \textbf{Techniques}  & \textbf{Methods} \\
			\hline
			\multirow{7}{1.8cm}{Input augmentations}
			& Additional Noise  & Ladder Network \cite{DBLP:conf/nips/RasmusBHVR15},  WCP \cite{DBLP:conf/cvpr/ZhangLH20WCP}\\
			\cline{2-3}
			& Stochastic Augmentation &$\Pi$ Model \cite{DBLP:conf/nips/SajjadiJT16}, Temporal Ensembling \cite{DBLP:conf/iclr/LaineA17}, Mean Teacher \cite{DBLP:conf/nips/TarvainenV17}, Dual Student \cite{DBLP:conf/iccv/KeWYRL19}, MixMatch \cite{DBLP:conf/nips/BerthelotCGPOR19}, ReMixMatch \cite{DBLP:conf/iclr/BerthelotCCKSZR20}, FixMatch \cite{DBLP:conf/nips/SohnBCZZRCKL20} \\
			\cline{2-3}
			& Adversarial perturbation  & VAT \cite{DBLP:journals/pami/MiyatoMKI19}, VAdD \cite{DBLP:conf/aaai/ParkPSM18} \\
			\cline{2-3}
			& AutoAugment  & UDA \cite{DBLP:conf/nips/XieDHL020}, Noisy Student \cite{DBLP:conf/cvpr/XieLHL20}  \\
			\cline{2-3}
			& RandAugment  & UDA \cite{DBLP:conf/nips/XieDHL020}, FixMatch \cite{DBLP:conf/nips/SohnBCZZRCKL20} \\
			\cline{2-3}
			& CTAugment  & ReMixMatch \cite{DBLP:conf/iclr/BerthelotCCKSZR20}, FixMatch \cite{DBLP:conf/nips/SohnBCZZRCKL20} \\
			\cline{2-3}
			& Mixup &ICT \cite{DBLP:conf/ijcai/VermaLKBL19}, MixMatch \cite{DBLP:conf/nips/BerthelotCGPOR19}, ReMixMatch \cite{DBLP:conf/iclr/BerthelotCCKSZR20}, DivideMix \cite{DBLP:conf/iclr/LiSH20}\\
			\hline
			\multirow{5}{1.8cm}{Neural network transformations}
			& Dropout  &$\Pi$ Model \cite{DBLP:conf/nips/SajjadiJT16}, Temporal Ensembling \cite{DBLP:conf/iclr/LaineA17},  Mean Teacher \cite{DBLP:conf/nips/TarvainenV17}, Dual Student \cite{DBLP:conf/iccv/KeWYRL19}, VAdD \cite{DBLP:conf/aaai/ParkPSM18},Noisy Student~\cite{DBLP:conf/cvpr/XieLHL20}  \\
			\cline{2-3}
			& EMA & Mean Teacher \cite{DBLP:conf/nips/TarvainenV17}, ICT \cite{DBLP:conf/ijcai/VermaLKBL19}  \\
			\cline{2-3}
			& SWA &SWA \cite{DBLP:conf/iclr/AthiwaratkunFIW19}  \\
			\cline{2-3}
			& Stochastic depth  & Noisy Student \cite{DBLP:conf/cvpr/XieLHL20}  \\
			\cline{2-3}
			& DropConnect &WCP \cite{DBLP:conf/cvpr/ZhangLH20WCP}  \\
			\hline
		\end{tabular}
	\end{table}

	\section{Challenges and future directions}\label{sec:future_trends}
	
	Although exceptional performance and achieved promising DSSL progress, there are still several open research questions for future work. We outline some of these issues and future directions below.
	
	\textbf{Theoretical analysis.} Existing semi-supervised approaches predominantly use unlabeled samples to generate constraints and then update the model with labeled data and these constraints. However, the internal mechanism of DSSL and  the role of various techniques, such as data  augmentations, training methods and loss functions,\etc, are not clear. 
	Generally, there is a single weight to balance the supervised and unsupervised loss, which means that all unlabeled instances are equally weighted. However, not all unlabeled data is equally appropriate for the model in practice. To counter this issue, \cite{DBLP:conf/nips/RenYS20} considers how to use a different weight for every unlabeled example. 
	%An algorithm adjusts those weights based on the influence function, which evaluates the dependence of the model on one training example. 
	For consistency regularization SSL, \cite{DBLP:conf/iclr/AthiwaratkunFIW19} investigates how loss geometry interacts with training process. \cite{DBLP:conf/nips/ZophGLCLC020} experimentally explores the effects of data augmentation and labeled dataset size on pre-training and self-training, as well as the limitations and interactions of pre-training and self-training. \cite{DBLP:conf/iclr/GhoshT21} analyzes the property of the consistency regularization methods when data instances lie in the neighborhood of low-dimensional manifolds, especially in the case of efficient data augmentations or perturbations schemes.
	
	\textbf{Incorporation of domain knowledge.} Most of the SSL approaches listed above can obtain satisfactory results only in ideal situations in which the training dataset meets the designed assumptions and contains sufficient information to learn an insightful learning system. However, in practice, the distribution of the dataset is unknown and does not necessarily meet these ideal conditions. When the distributions of labeled and unlabeled data do not belong to the same one or the model assumptions are incorrect, the more unlabeled data is utilized, the worse the performance will be. Therefore, we can attempt to incorporate richer and more reliable domain knowledge into the model to mitigate the degradation performance. Recent works focusing on this direction have proposed \cite{DBLP:conf/acl/HuMLHX16,DBLP:journals/pami/TangWWGDGC18,DBLP:conf/cvpr/QiWQL19,DBLP:journals/mlc/YuYZ19, DBLP:journals/corr/abs-2008-03923} for DSSL.
	
	\textbf{Learning with noisy labels.} In this survey, models discussed typically consider the labeled data is generally accurate to learn a standard cross-entropy loss function. An interesting consideration is to explore how SSL can be performed for cases where corresponding labeled instances with noisy initial labels.
	For example, the labeling of samples  may be contributed by the community, so we can only obtain noisy labels for the training dataset. One solution to this problem is \cite{DBLP:journals/corr/ReedLASER14}, which augments the prediction objective with consistency where the same prediction is made given similar percepts. Based on graph SSL, \cite{DBLP:journals/pr/LuW15a}  introduces a new $L_1$-norm formulation of Laplacian regularization inspired by sparse coding. \cite{DBLP:conf/nips/HanYYNXHTS18} deals with this problem from the perspective of memorization effects, which proposed a learning paradigm combining co-training and mean teacher.
	
	\textbf{Imbalanced semi-supervised learning.} The problem of class imbalance is naturally widespread in real-world applications. When the training data is highly imbalanced, most learning frameworks will show bias towards the majority class, and in some extreme cases, may completely ignore the minority class \cite{DBLP:journals/jbd/JohnsonK19}, as a result, the efficiency of predictive  models will be significantly affected. Nevertheless, to handle the semi-supervised problem, it is commonly assumed that training dataset is uniformly distributed over all class labels. Recently, more and more works have focused on this problem. \cite{DBLP:conf/nips/KimHPYHS20} aligns pseudo-labels with the desirable class distribution in the unlabeled data for SSL with imbalanced labeled data. Based on graph-based semi-supervised  methods, \cite{DBLP:journals/pr/DengY21} copes with various  degrees of class imbalance in a given dataset.
	
	%distribution-aware semantics-oriented pseudo-label for imbalanced semi-supervised learning \cite{DBLP:journals/corr/abs-2106-05682}
	
	%Rethinking re-sampling in imbalanced semi-supervised learning. \cite{DBLP:journals/corr/abs-2106-00209}
	
	%Semi-supervised long-tailed recognition using alternate sampling.\cite{DBLP:journals/corr/abs-2105-00133}

	\textbf{Robust semi-supervised learning.} The common of the latest state-of-the-art  approaches  is the application of consistency training on augmented unlabeled data without changing the model predictions. One attempt is made by VAT \cite{DBLP:journals/pami/MiyatoMKI19} and VAdD \cite{DBLP:conf/aaai/ParkPSM18}. Both of them employ adversarial training to find the optimal adversarial example. Another promising approach is data augmentation (adding noise or random perturbations, CutOut \cite{DBLP:journals/corr/abs-1708-04552}, RandomErasing \cite{DBLP:conf/aaai/Zhong0KL020}, HideAndSeek \cite{DBLP:journals/corr/abs-1811-02545} and GridMask \cite{DBLP:journals/corr/abs-2001-04086}), especially advanced data augmentation, such as AutoAugment \cite{DBLP:conf/cvpr/CubukZMVL19}, RandAugment \cite{DBLP:conf/cvpr/CubukZSL20}, CTAugment \cite{DBLP:conf/iclr/BerthelotCCKSZR20}, and Mixup \cite{DBLP:conf/iclr/ZhangCDL18} which also  can be considered as a form of regularization.

	\textbf{Safe semi-supervised learning.} In SSL, it is generally accepted that unlabeled data can help improve learning performance, especially when labeled data is scarce. Although it is remarkable that unlabeled data can improve learning performance under appropriate assumptions or conditions, some empirical studies \cite{DBLP:conf/nips/SinghNZ08,DBLP:journals/pami/YangP11,DBLP:journals/jair/ChawlaK05} have shown that  the use of unlabeled data may lead to performance degeneration, making the generalization performance even worse than a model learned only with labeled data in real-world applications. Thus, safe semi-supervised learning approaches are desired, which never significantly degrades learning performance when unlabeled data is used.

	\section{Conclusion}\label{sec:conclusion}
	Deep semi-supervised learning is a promising research field with important real-world applications. The success of deep learning approaches has led to the rapid growth of DSSL techniques. This survey provides a taxonomy of existing DSSL methods, and groups DSSL methods into five categories: Generative models, Consistency Regularization models, Graph-based models, Pseudo-labeling models, and Hybrid models. We provide illustrative figures to compare the differences between the approaches within the same category. Finally, we discuss the challenges of DSSL and some future research directions that are worth further studies.

	\ifCLASSOPTIONcompsoc
	%\else
	%  % regular IEEE prefers the singular form
	\section*{Acknowledgment}
	\fi
	This paper was partially supported by the National Key Research and Development Program of China (No. 2018AAA0100204), and a key  program of fundamental research from Shenzhen Science and Technology Innovation Commission (No. JCYJ20200109113403826).
	

	\ifCLASSOPTIONcaptionsoff
	\newpage
	\fi

	
	
	%\bibliographystyle{IEEEtran}
	\bibliographystyle{IEEEtran}
	\bibliography{IEEEabrv,biblio}
	
	%\newpage

	%\newpage
	
	
	
	%\vfill
	
	
\end{document}